\documentclass{article}

\usepackage{arxiv}

\usepackage[utf8]{inputenc} 
\usepackage[T1]{fontenc}    
\usepackage{hyperref}       
\usepackage{url}            
\usepackage{booktabs}       
\usepackage{amsfonts}       
\usepackage{nicefrac}       
\usepackage{microtype}      
\usepackage{lipsum}         
\usepackage{graphicx}
\usepackage[numbers]{natbib}
\usepackage{doi}

\usepackage[table,xcdraw]{xcolor}
\usepackage{subfigure}
\usepackage[normalem]{ulem}
\usepackage{array}
\usepackage{caption} 
 \usepackage{multirow} 
 \usepackage{amsmath}
\usepackage{amssymb}
\usepackage{mathtools}
\usepackage{amsthm}
\usepackage{cleveref}       

\usepackage{bm}

\title{Decomposing the Time Series Forecasting Pipeline: \\ A Modular Approach for Time Series Representation, Information Extraction, and Projection}

\date{}

\newif\ifuniqueAffiliation

\ifuniqueAffiliation 

\else
\usepackage{authblk}

\newbox{\orcid}\sbox{\orcid}{} 
\author[1]{%
	{\usebox{\orcid}\hspace{1mm}Robert Leppich}%
}
\author[1]{%
	{\usebox{\orcid}\hspace{1mm}Michael Stenger}%
}
\author[2]{%
	{\usebox{\orcid}\hspace{1mm}André Bauer}%
}
\author[1]{%
	{\usebox{\orcid}\hspace{1mm}Samuel Kounev}%
}
\affil[1]{Department of Computer Science, University of Wuerzburg, Germany}
\affil[2]{Illinois Institute of Technology, Chicago, US}

\fi

\renewcommand{\headeright}{PREPRINT}
\renewcommand{\shorttitle}{RepNet}

\hypersetup{
pdftitle={Decomposing the Time Series Forecasting Pipeline: A Modular Approach for Time Series Representation, Information Extraction, and Projection},
}

\def\eqref#1{equation~\ref{#1}}

\def\1{\bm{1}}

\def\vx{{\bm{x}}}
\def\vy{{\bm{y}}}

\DeclareMathAlphabet{\mathsfit}{\encodingdefault}{\sfdefault}{m}{sl}
\SetMathAlphabet{\mathsfit}{bold}{\encodingdefault}{\sfdefault}{bx}{n}

\begin{document}
\maketitle

\begin{abstract}
With the advent of Transformers, time series forecasting has seen significant advances, yet it remains challenging due to the need for effective sequence representation, memory construction, and accurate target projection.
Time series forecasting remains a challenging task, demanding effective sequence representation, meaningful information extraction, and precise future projection. 
Each dataset and forecasting configuration constitutes a distinct task, each posing unique challenges the model must overcome to produce accurate predictions. To systematically address these task-specific difficulties, this work decomposes the time series forecasting pipeline into three core stages: input sequence representation, information extraction and memory construction, and final target projection. Within each stage, we investigate a range of architectural configurations to assess the effectiveness of various modules, such as convolutional layers for feature extraction and self-attention mechanisms for information extraction, across diverse forecasting tasks, including evaluations on seven benchmark datasets.
Our models achieve state-of-the-art forecasting accuracy while greatly enhancing computational efficiency, with reduced training and inference times and a lower parameter count. 
The source code is available at \url{https://github.com/RobertLeppich/REP-Net}.
\end{abstract}


\section{Introduction}
\label{sec:intro}

Time series analysis has high potential in both science and industry. It comprises various disciplines, including time series forecasting~\cite{forecasting}, classification~\cite{classification}, and imputation~\cite{imputation}. By analyzing time series data, we can gain deeper insights into various systems, such as sensor networks~\citep{papadimitriou2006optimal}, finance~\citep{zhu2002statstream}, and biological systems like the human body~\citep{ek2023transformer}. Among these tasks, time series forecasting plays a vital role in predicting future trends and supporting data-driven decision-making across domains such as finance, healthcare, and supply chain management, facilitating improved resource allocation and risk mitigation. In contrast, time series imputation is essential for handling missing data, particularly in sensor-generated time series, ensuring data integrity and reliability for downstream analysis.

Time series (TS) data are often high-dimensional, characterized by complex relationships shaped by both temporal dependencies and attribute-level structures. Typically recorded as continuous streams, time series capture one or more values at each time step. However, single observations usually carry limited semantic information, which leads to a research focus on temporal variations. These variations reveal essential properties of TS data, such as continuity and intricate temporal patterns. Modeling such dynamics is particularly challenging, as multiple overlapping temporal variations may coexist, adding layers of complexity to the analysis.


Despite significant methodological advances---such as Recurrent Neural Networks~(RNNs) and Convolutional Neural Networks~(CNNs)---challenges like the curse of dimensionality and vanishing/exploding gradients persist, restricting the information flow over long sequences~\cite{hochreiter2001gradient}.
The emergence of the Transformer architecture~\cite{vaswani2017attention}, initially in Natural Language Processing (NLP) and later adopted in time series (TS) analysis~\citep{wu2020deep}, helped address modeling of termporal variations. However, due to its point-wise attention mechanism, standard Transformer-based models struggled to fully capture essential TS characteristics and faced scalability issues on long sequences~\citep{huang2018improved, povey2018time}. In response, more specialized models have been developed, incorporating mechanisms such as sparse, frequency-based, and cross-variable attention~~\cite{zhou2021informer, wu2021autoformer, zhou2022fedformer, zhang2023crossformer}. Other approaches focus on input segmentation, for instance, with patching~\citep{nie2022time, liu2024itransformer, chen2024pathformer} or hierarchical representations of TS~\citep{wu2022timesnet}. In a different line of research, several works focused on simpler, more light-wight architectures revolving around linear layers and the multi-layer perceptron (MLP)~\cite{zeng2023transformers, li2023revisiting, chen2023tsmixer, lin2024cyclenet, hu2024timecnn}. More recently, researchers have also investigated the use of novel state space models for TS forecasting~\cite{gu2021efficiently,ahamed2024timemachine}, as well as the development of time series foundation models, which have shown strong performance not only in forecasting but also across a range of TS tasks~\cite{zhang2022self, wu2022timesnet, zhou2023one, rasul2023lag, goswami2024moment, woo2024unified, das2024decoder}.
We provide a extended related work section and delimitation to our work in Appendix~\ref{appendix:related_work}.

The diversity of forecasting tasks, characterized by inherent attributes of the datasets and varying forecasting horizons, introduces distinct challenges that influence the design requirements of forecasting models.
Some datasets exhibit simple periodic patterns with minimal temporal variation, while others contain more complex structures that are, in some cases, closely tied to the underlying timeline. 
The forecasting horizon defines the temporal scale of patterns the model must learn to produce accurate forecasts. Short horizons primarily demand the capture of fine-grained, short-term patterns, while longer horizons require modeling of seasonalities and long-term trends.

In this work, we decompose the time series forecasting pipeline to address the varying requirements of different forecasting tasks and to analyze the impact of architectural variations on forecasting performance across diverse scenarios.

To identify optimal configurations for time series forecasting tasks, we propose REP-Net, a novel architecture that divides the forecasting process into three key modules: \textbf{R}epresentation of the input sequence (Representation), \textbf{E}nrichment of information into a memory module (Memory), and \textbf{P}rojection of the memory onto the target output space (Projection).

For each module, we explore established architectural designs, such as self-attention, and introduce novel components, including time-informed patches and patch-based temporal pattern modeling, to identify key factors contributing to high-quality forecasting performance for each task.

We summarize the key contributions of this work as follows:

\begin{itemize}
    \item We propose REP-Net, a novel architecture that decomposes the time series forecasting process into three distinct modules: Representation, Memory, and Projection. Each module is designed to flexibly incorporate a wide range of established and novel techniques for time series forecasting.
    \item We conduct a comprehensive empirical evaluation on seven established benchmark datasets for time series forecasting, comparing REP-Net with state-of-the-art (SOTA) models. REP-Net consistently matches or surpasses SOTA performance, demonstrating robustness and efficiency across diverse data distributions.
    \item We provide a detailed analysis of architectural variations and their influence on the TS forecasting quality. 
    \item To foster reproducibility and further research, we publicly release the REP-Net codebase, enabling researchers to benchmark and extend our work.
\end{itemize}

\section{Methodology}\label{sec:methodology}
In the following, we introduce REP-Net, a novel and highly flexible architecture tailored to meet the specific requirements of a wide range of time series (TS) forecasting tasks. We define TS forecasting as follows:
Given multivariate time series instances $\vx_1 \dots \vx_T$, where $\vx_i \in \mathbb{R}^F$, $F$ is the number of input features, and $T$ is the input length, the objection is to forecast $\hat{\vy}_1 \dots \hat{\vy}_H$, where $\hat{\vy}_i \in \mathbb{R}^F$ and $H$ is the forecasting horizon.

\subsection{Architecture}

\begin{figure*}[t]
\centering
\includegraphics[width=.9\textwidth]{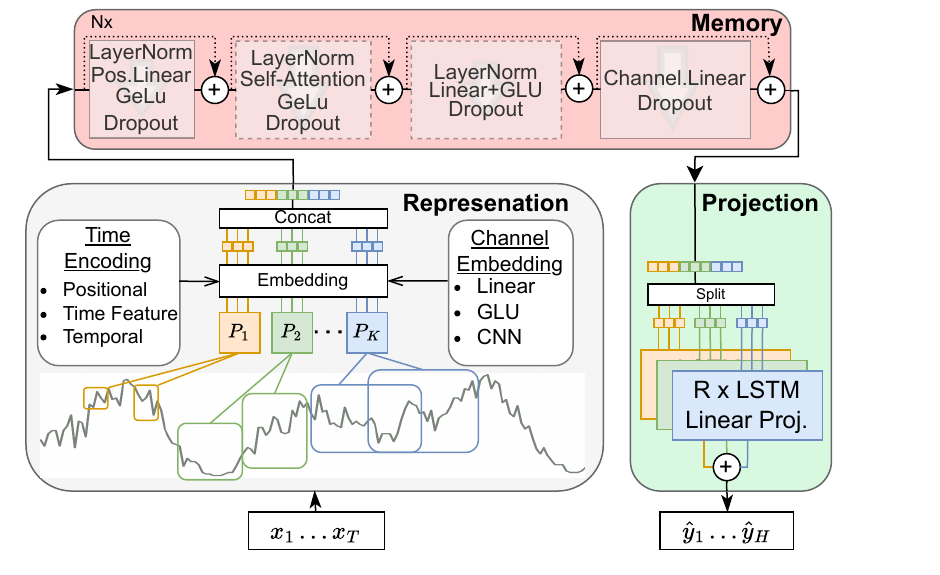}

\caption{Illustraton of the proposed REP-Net Architecture, consisting of three modules: Representation (gray), Memory (red), and Projection (green).}
\label{figure:app:architecture_all}
\end{figure*}

As described above and illustrated in Figure~\ref{figure:app:architecture_all}, we strictly structure the forecasting process into three distinct phases: (i) Representation, (ii) Memory, and (iii) Projection. 
%
The input sequence $\vx_1, \dots, \vx_T$ is processed by the representation module, shown in gray in Figure~\ref{figure:app:architecture_all}, where it is segmented into diverse patches using $K$ independent patch extractors. Each resulting patch includes an abstraction of the input along with its corresponding temporal information. In parallel, temporal information is embedded using the same embedding method as for the features, but with separate gradients.
The feature and temporal embeddings are concatenated to form a unified patch representation, called time-informed patch. All patch representations are subsequently concatenated into a single matrix, enabling the memory module in the next stage to access information from all patches across different abstraction levels simultaneously.
The memory module, shown in red in Figure~\ref{figure:app:architecture_all}, comprises up to four layer blocks, depending on the specific configuration. Each block includes residual connections to promote gradient flow. The memory module is stacked $N$ times, and its output is passed to the final component, the projection module.
The projection module, illustrated in green in Figure~\ref{figure:app:architecture_all}, comprises $R$ LSTM layers followed by a final linear projection layer. The input is split according to the concatenation from the output of the representation layer. Each of the $K$ splits is processed independently using separate LSTM and projection layers. The outputs of the $K$ projections are summed to produce the final forecast.
In the following, we describe the three modules in detail.

\subsection{Representation}\label{sec:representation}
The representation of the input sequence is a critical step, as it must effectively capture and preserve all relevant information, including structural and temporal dependencies, in a format suitable for downstream processing. Given the inherently high dimensionality of time series data, as previously discussed, this information must be compressed into a lower-dimensional representation to serve as an appropriate encoding for subsequent stages. A key objective of this step is to retain essential patterns within the input sequence, enabling the model to leverage them during forecasting.

Prior work has explored various representation methods to represent input sequences effectively. Recurrent-based architectures, such as~\citep{jhin2024addressing}, leverage the temporal modeling capabilities of recurrent units like GRUs and LSTMs to capture sequential dependencies. To enhance pattern recognition, some approaches, such as \citep{hu2024timecnn}, adopt convolutional neural network (CNN)–based representations, while others combine recurrent and convolutional techniques to exploit both temporal and spatial features.
Leppich et al.~\citep{leppich2025tsrm} proposed a CNN-based representation layer that independently learns features at multiple levels of abstraction from an input sequence. This layer consists of several CNN modules with varying configurations, including CNNs with small kernels to capture fine-grained patterns and CNNs with larger kernels and high dilation rates to model higher-level structures such as trends. The outputs of all CNN modules are concatenated, enabling subsequent stages to access information from all abstraction levels simultaneously.
Other approaches employ functions, such as fast Fourier transformation, to determine different abstraction levels of the time series~\citep{wu2022timesnet, chen2024pathformer}.
Another approach involves dividing the time series into smaller segments, referred to as patches. Each patch is represented as a vector, which reduces complexity and enables the model to process meaningful chunks of the time series rather than individual time steps~\citep{nie2022time, liu2024itransformer}.

In this work, we investigate various techniques for representing time series data through patch-based representations. Our method enables the extraction of patches from the input sequence at multiple levels of abstraction and sampling density, following a strategy similar to that proposed by Leppich et al.~\cite{leppich2025tsrm}. Specifically, we employ $K$ distinct patch extractors $P$, each configured with a unique combination of cover size, dilation, and stride. As illustrated in Figure~\ref{figure:app:architecture_all}, the yellow patch extractor $P_1$ utilizes small cover sizes, no dilation, and large strides, effectively capturing fine-grained local patterns in the time series. In contrast, the blue patch extractor $P_K$ employs larger cover sizes and high dilation rates to extract coarse, high-dimensional patterns such as long-term trends. In our experiments we evaluate $K \in [1, \dots, 5]$.

The sub-sequences extracted by each patch extractor $P$ are subsequently embedded using multiple embedding strategies $E$ to assess the effectiveness of each method across different forecasting horizons and datasets. In particular, we evaluate the following embedding strategies:

\begin{itemize}
    \item $E_1$: Embedding using linear layers.  
    \item $E_2$: Embedding using linear layers in combination with gated linear units (GLU).  
    \item $E_3$: Embedding using CNNs.
\end{itemize}

For each embedding strategy, we evaluate several variations, detailed in Appendix~\ref{app:representations}.

In addition to extracting sub-sequences from the input time series, the patch extractors also operate on the corresponding timeline, capturing features such as day of the week and hour of the day, using different time encoding. These sub-timelines are embedded using the same embedding method as the time series values, but with separate gradient flows during training.
The resulting temporal embeddings are then vertically concatenated with the time series embeddings to form time-informed patches. These patches jointly represent both an abstracted view of the input sequence and a corresponding abstraction of the temporal context, enabling the model to access both types of information simultaneously.

The resulting patches are concatenated horizontal into a single matrix, enabling the subsequent memory modules to access embeddings from all $K$ levels of abstraction of both, the TS and temporal information.

\subsection{Memory}\label{sec:memory}
After converting the input sequence into concatenated time-informed patches, the memory module extracts essential information from this representation into a dedicated memory. Identifying key patterns across the input sequence is critical at this stage, enabling the final module to generate high-quality forecasts.

Previous works have employed various strategies to address this problem. With the emergence of the Transformer architecture by Vaswani et al.~\cite{vaswani2017attention}, many time series forecasting models have incorporated components of the Transformer, specifically the encoder and decoder modules, to leverage the attention mechanism's ability to dynamically model dependencies across the entire input sequence~\cite{nie2022time, wu2021autoformer, zhou2022fedformer}.
Recent approaches build upon the work of Tolstikhin et al.~\cite{tolstikhin2021mlp}, which relies exclusively on linear layers and dimensional transformations to achieve high-quality results in computer vision. This concept has since been adapted to the time series domain by methods such as those proposed by Chen et al.~\cite{chen2023tsmixer}.

Our information extraction and enrichment module builds upon the time/feature MLP-mixing framework introduced by Chen et al.~\cite{chen2023tsmixer}. We extend this design by incorporating normalization, activation, and dropout layers. The complete module architecture is depicted in Figure~\ref{figure:app:architecture_all} (top).
The module comprises four sequential layer blocks. The first block contains a positional linear transformation, while the last block applies a feature-level linear transformation. The feature-level linear transformation block can be configured to operate either across all feature embeddings, enabling the memory modules to capture inter-feature dependencies, or on each feature embedding individually. An optional self-attention block is inserted between these two blocks, enabling the model to capture temporal dependencies when enabled. As attention function we employ sparse self-attention inspired by Wu et al.~\cite{wu2020adversarial}.
After the attention block, the memory module includes additional feature-level linear layer followed by a gated linear unit (GLU), which facilitates selective information retention by allowing the model to discard irrelevant features.
All blocks and the entire module are equipped with residual connections to facilitate information and gradient flow.
The module is stacked $N$ times, enabling deep architectures to capture hierarchical information and patterns.

In our experiments, we assess the impact of the attention block and the GLU block on model performance. Additionally, we vary the number of stacked memory modules ($N$) to evaluate their effect across different datasets and forecasting horizons.

\subsection{Projection}\label{sec:projection}
The final step in the time series forecasting process involves leveraging the information from the previous stage to generate predictions over a forecasting horizon, denoted as $\hat{\vy}_1, \dots, \hat{\vy}_H$. Recurrent approaches typically adopt an autoregressive strategy, using encoded information from earlier time steps to predict the next one~\citep{jhin2024addressing}. Similarly, transformer-based methods often employ an autoregressive framework by applying causal masks, which ensure that the prediction at time step $t$ depends only on inputs from time steps $\le t$~\citep{zhou2022fedformer, wu2021autoformer}.
Recent approaches have shown that relatively simple architectures leveraging time series decomposition, combined with either linear layers or a single recurrent layer followed by a linear layer, can achieve competitive performance with low computational overhead. These models often outperform or match certain Transformer-based architectures, depending on the benchmark dataset and forecasting horizon~\cite{li2023revisiting, zeng2022transformers}. Recognizing the effectiveness of linear layers for projections, many methods have adopted this design choice with notable success~\cite{nie2022time, leppich2025tsrm}.

The projection module, shown in Figure~\ref{figure:app:architecture_all} (green), comprises $K$ separate projection blocks, where $K$ corresponds to the number of patch extractors in the Representation module. Before entering the projection module, the concatenated patches are split back into their original partitions as produced by the patch extractors. An individual projection block then processes each of the resulting $K$ patch sequences. The outputs of these $K$ projection blocks are added to produce the final forecasting result.

We investigate various configurations of the projection block, each comprising $R$ LSTM layers followed by a linear layer. The LSTM layers are intended to capture temporal dependencies within each patch sequence, thereby enhancing the forecasting capability of the subsequent linear layer. We evaluate the effect of varying the number of LSTM layers with $R \in {0, 1, 2, 3, 4}$, where $R = 0$ indicates the absence of LSTM layers. Our experiments assess the impact of these configurations on forecasting performance across different benchmark datasets and forecast horizons.

We provide a detailed hyperparameter setup for each module in the Appendix~\ref{app:hyperparameters}.

\section{Experiments}\label{sec:experiments}

In order to assess the effectiveness of our proposed architecture, we conducted a series of experiments using publicly accessible and well-established benchmark datasets from different fields: ECL~\citep{dua2017uci}, ETT~\citep{zhou2021informer} (four subsets: ETTm1, ETTm2, ETTh1, ETTh2), Weather~\citep{wu2021autoformer}, and Traffic~\cite{traffic_dataset}. All datasets were collected from~\cite{wu2021autoformer}. For more details, we refer to Section~\ref{appendix:datasets}.

\subsection{Experimental Setup}\label{sec:exp_setup}

For all experiments, we applied early stopping with a patience of three epochs, using the validation loss as the evaluation metric. The training was terminated when the relative improvement of the loss was less than 1\%. 
Hyperparameter tuning was performed via random search over the following configuration space. 

For the representation module, we explored various patch extractors $P$ and representation techniques as described in Section~\ref{sec:representation}. Additionally, we investigate different encoding sizes and time embedding techniques.
For the memory module, we vary the number of memory modules (denoted by $N$), the inclusion of the attention block, and the presence of the GLU block, as described in Section~\ref{sec:memory}. We also explore different numbers of attention heads in the self-attention layer. Additionally, we examine whether the first feature-level linear layer in the memory modules operates across embeddings of all features jointly or remains feature-agnostic.
For the Projection module, we vary the number of LSTM layers.
A comprehensive list of all hyperparameters is provided in Appendix~\ref{app:hyperparameters}.

Learning rates were dynamically adapted during training using a learning rate scheduler (ReduceLROnPlateau) from \textit{PyTorch} with a patience of one epoch. All models were trained with the Adam optimizer on an Nvidia A100 80GB GPU.

\subsection{Long-Term Time Series Forecasting}\label{sec:exp:fc}

To measure the discrepancy between the prediction sequence $\hat{\vy}_{1} \dots \hat{\vy}_{H}$ and the ground truth $\vy_{1} \dots \vy_{H}$ for a horizon $H$, where $\hat{\vy}_{i}, \vy_{i} \in \mathbb{R}^F$, we used the Huber loss function during training. 
The loss is calculated for each element pair $\hat{y}_{ij} \in \mathbb{R}, y_{ij} \in \mathbb{R}$ and averaged across all $F$ channels and $H$ timesteps to obtain the overall objective loss. Parameter $\delta > 0$ depends on the dataset.

\begin{equation}
    \label{eq:huber_loss}
    \mathcal{L}_\text{Huber} = \frac{1}{FH} \sum^{H}_{i=1} \sum^{F}_{j=1} l_{ij},
    \qquad
    l_{ij} = 
    \begin{cases}
			\frac{1}{2}(\hat{y}_{ij} - y_{ij})^2, & \text{if } |\hat{y}_{ij} - y_{ij}| < \delta \\
            \delta \cdot (|\hat{y}_{i} - y_{i}| - \frac{1}{2}\delta), & \text{otherwise.}
	\end{cases}
\end{equation}

To evaluate the performance of our architecture for long-term TS forecasting, we adopted the procedure by~\cite{liu2024itransformer}:
To support a fair comparison with other approaches, we maintain the input length for all approaches at $T=96$, while varying the prediction horizon $H \in \{96, 192, 336, 720\}$. For the evaluation, we consider seven datasets (i.e., ECL, ETTm1, ETTm2, ETTh1, ETTh2, Weather, and Traffic) and compare against multiple SOTA forecasting techniques: CycleNet~\cite{lin2024cyclenet}, TimeMixer~\cite{wang2024timemixer}, TimeMachine~\cite{timemachine}, TimeCNN~\cite{hu2024timecnn}, TSRM~\cite{leppich2025tsrm}, PathFormer~\cite{chen2024pathformer},  iTransformer~\citep{liu2024itransformer}, and PatchTST~\citep{nie2022time}.

\subsection{Results}
Table~\ref{tab:results} reports the overall best results across all module configurations compared to SOTA architectures. REP-Net consistently delivers strong performance across most datasets and prediction lengths, demonstrating superior forecasting accuracy, particularly at the longest horizon of 720 (see Appendix~\ref{app:tab:results} for detailed results). For shorter horizons, our method largely matches SOTA performance, suggesting that the predictive capacity of the datasets is approaching saturation and that performance metrics are beginning to converge.
However, our approach did not outperform SOTA methods on the Traffic dataset. Further analysis revealed that feature 840 contained significant outliers in the test set, substantially increasing the MSE metric. After excluding this feature, our model achieved an average MSE of $0.372$ and an average MAE of $0.279$ over all four prediction length.

\begin{table}[]
\centering
    \caption{Performance comparison for the multivariate forecasting task with prediction horizons $H \in \{96, 192, 336, 720\}$ and fixed lookback window $T = 96$. Results are averaged over all prediction horizons. Bold/underline indicate best/second. For the best results, we include a margin of 1\% to be marked bold as well. }
    \label{tab:results}
\scalebox{.7}{
\begin{tabular}{@{}cllllllllllllllllll@{}}
\toprule
Dataset & \multicolumn{2}{c}{\textbf{REP-Net}} & \multicolumn{2}{c}{\begin{tabular}[c]{@{}c@{}}CycleNet \\ (NeurIPS 2024)\end{tabular}} & \multicolumn{2}{c}{\begin{tabular}[c]{@{}c@{}}TimeMixer \\ (ICLR 2024)\end{tabular}} & \multicolumn{2}{c}{\begin{tabular}[c]{@{}c@{}}TimeMachine \\ (ECAI 2024)\end{tabular}} & \multicolumn{2}{c}{TimeCNN} & \multicolumn{2}{c}{TSRM} & \multicolumn{2}{c}{\begin{tabular}[c]{@{}c@{}}PathFormer\\ (ICLR 2024)\end{tabular}} & \multicolumn{2}{c}{\begin{tabular}[c]{@{}c@{}}iTransformer\\ (ICLR 2024)\end{tabular}} & \multicolumn{2}{c}{\begin{tabular}[c]{@{}c@{}}PatchTST\\ (ICLR 2023)\end{tabular}} \\ \midrule
\multicolumn{1}{c|}{} & \multicolumn{1}{c}{MSE} & \multicolumn{1}{c|}{MAE} & \multicolumn{1}{c}{MSE} & \multicolumn{1}{c}{MAE} & \multicolumn{1}{c}{MSE} & \multicolumn{1}{c}{MAE} & \multicolumn{1}{c}{MSE} & \multicolumn{1}{c}{MAE} & \multicolumn{1}{c}{MSE} & \multicolumn{1}{c}{MAE} & \multicolumn{1}{c}{MSE} & \multicolumn{1}{c}{MAE} & \multicolumn{1}{c}{MSE} & \multicolumn{1}{c}{MAE} & \multicolumn{1}{c}{MSE} & \multicolumn{1}{c}{MAE} & \multicolumn{1}{c}{MSE} & \multicolumn{1}{c}{MAE} \\
\multicolumn{1}{c|}{ECL} & \textbf{0.165} & \multicolumn{1}{l|}{0.266} & \underline{0.167} & \textbf{0.258} & 0.182 & 0.273 & 0.170 & \underline{0.263} & 0.170 & 0.264 & 0.175 & 0.270 & 0.182 & 0.269 & 0.178 & 0.270 & 0.190 & 0.275 \\
\multicolumn{1}{c|}{Weather} & \textbf{0.234} & \multicolumn{1}{l|}{\textbf{0.263}} & 0.243 & 0.271 & 0.240 & 0.272 & 0.243 & 0.273 & 0.252 & 0.278 & \underline{0.239} & 0.267 & \underline{0.239} & \textbf{0.263} & 0.258 & 0.278 & 0.261 & 0.285 \\
\multicolumn{1}{c|}{Traffic} & 0.437 & \multicolumn{1}{l|}{\underline{0.279}} & 0.472 & 0.301 & 0.485 & 0.298 & 0.429 & 0.281 & \textbf{0.410} & \textbf{0.273} & 0.545 & 0.324 & 0.501 & 0.299 & \underline{0.428} & 0.282 & 0.467 & 0.293 \\
\multicolumn{1}{c|}{ETTh1} & \textbf{0.417} & \multicolumn{1}{l|}{\underline{0.425}} & 0.432 & 0.428 & 0.447 & 0.440 & \textbf{0.417} & \textbf{0.419} & 0.441 & 0.436 & 0.435 & 0.430 & 0.439 & 0.430 & 0.454 & 0.448 & 0.455 & 0.444 \\
\multicolumn{1}{c|}{ETTh2} & 0.352 & \multicolumn{1}{l|}{0.389} & 0.382 & 0.406 & 0.365 & 0.395 & \textbf{0.344} & \textbf{0.379} & 0.376 & 0.402 & 0.371 & 0.397 & \textbf{0.344} & \textbf{0.379} & 0.383 & 0.407 & 0.378 & 0.402 \\
\multicolumn{1}{c|}{ETTm1} & \textbf{0.366} & \multicolumn{1}{l|}{\underline{0.390}} & 0.379 & 0.396 & 0.381 & 0.396 & \underline{0.375} & 0.392 & 0.383 & 0.392 & 0.380 & 0.393 & 0.382 & \textbf{0.386} & 0.407 & 0.410 & 0.384 & 0.396 \\
\multicolumn{1}{c|}{ETTm2} & \underline{0.264} & \multicolumn{1}{l|}{\textbf{0.314}} & 0.266 & \textbf{0.314} & 0.275 & 0.323 & 0.268 & 0.318 & 0.280 & 0.325 & \textbf{0.245} & \textbf{0.314} & 0.273 & \textbf{0.314} & 0.290 & 0.332 & 0.284 & 0.327 \\ \bottomrule
\end{tabular}
}
\end{table}

Our proposed architecture exhibits a stable performance that meets or even exceeds SOTA results. However, in addition to the pure performance metrics, a model's complexity and runtime should also play an essential role when assessing an architecture's quality, not least in order to ensure its cost efficiency and applicability in practice. 
REP-Net is designed for memory efficiency with a small memory footprint.
Table~\ref{tab:efficientcy} presents a comparison of trainable parameters, inference time per iteration, and GPU memory usage across four datasets for four SOTA architectures. The final row reports the average values computed across all datasets. All experiments were conducted using a fixed input length and forecasting horizon of $H = T = 96$, with a batch size of one.
Our proposed REP-Net architecture demonstrates solid efficiency in both inference time and memory footprint across all four datasets. In terms of trainable parameters, REP-Net ranks third among the evaluated models.

\begin{table}[]
\centering
\caption{Model efficiency comparison across four datasets. We report the number of trainable parameters (\textit{Params.}), inference time per iteration (\textit{Time}), and GPU memory consumption (\textit{Memory}). The last section shows the average across all datasets. All experiments were conducted using a fixed input length and forecasting horizon of $H=T=96$, with the batch size set to one.}
\label{tab:efficientcy}
\scalebox{.65}{
\begin{tabular}{@{}cccccccccccccccc@{}}
\toprule
Dataset & \multicolumn{3}{c}{ETTh1} & \multicolumn{3}{c}{ETTm1} & \multicolumn{3}{c}{ECL} & \multicolumn{3}{c}{Weather} & \multicolumn{3}{c}{Average} \\ \midrule
\multicolumn{1}{c|}{Models} & \begin{tabular}[c]{@{}c@{}}Param.\\ (K)\end{tabular} & \begin{tabular}[c]{@{}c@{}}Time\\ (s / iter)\end{tabular} & \begin{tabular}[c]{@{}c@{}}Memory\\ (MB)\end{tabular} & \begin{tabular}[c]{@{}c@{}}Param.\\ (K)\end{tabular} & \begin{tabular}[c]{@{}c@{}}Time\\ (s / iter)\end{tabular} & \begin{tabular}[c]{@{}c@{}}Memory\\ (MB)\end{tabular} & \begin{tabular}[c]{@{}c@{}}Param.\\ (K)\end{tabular} & \begin{tabular}[c]{@{}c@{}}Time\\ (s / iter)\end{tabular} & \begin{tabular}[c]{@{}c@{}}Memory\\ (MB)\end{tabular} & \begin{tabular}[c]{@{}c@{}}Param.\\ (K)\end{tabular} & \begin{tabular}[c]{@{}c@{}}Time\\ (s / iter)\end{tabular} & \begin{tabular}[c]{@{}c@{}}Memory\\ (MB)\end{tabular} & \begin{tabular}[c]{@{}c@{}}Param.\\ (K)\end{tabular} & \begin{tabular}[c]{@{}c@{}}Time\\ (s / iter)\end{tabular} & \begin{tabular}[c]{@{}c@{}}Memory\\ (MB)\end{tabular} \\
\multicolumn{1}{c|}{CycleNet} & 99 & 0.009 & 24 & 99 & 0.008 & 24 & 152 & 0.011 & 28 & 101 & 0.008 & 24 & 113 & 0.009 & 25 \\
\multicolumn{1}{c|}{TimeMixer} & 75 & 0.015 & 26 & 75 & 0.013 & 26 & 106 & 0.017 & 166 & 104 & 0.015 & 32 & 90 & 0.015 & 63 \\
\multicolumn{1}{c|}{Pathformer} & 438 & 0.134 & 34 & 681 & 0.130 & 44 & 3900 & 0.120 & 1280 & 728 & 0.110 & 74 & 1437 & 0.124 & 358 \\
\multicolumn{1}{c|}{TSRM} & 857 & 0.0120 & 30 & 1650 & 0.0100 & 31 & 161 & 0.019 & 110 & 338 & 0.008 & 32 & 752 & 0.012 & 51 \\
\multicolumn{1}{c|}{REP-Net} & 338 & 0.009 & 28 & 427 & 0.007 & 28 & 243 & 0.011 & 90 & 381 & 0.006 & 28 & 347 & 0.009 & 44 \\ \bottomrule
\end{tabular}
}
\end{table}

\section{Model Analysis}
As demonstrated above, our proposed architecture achieves competitive performance while maintaining a low memory footprint and fast computation, compared to SOTA architectures, when each of the three modules is configured with the optimal setting for the target task.
To assess the impact of individual architectural modifications, their interdependencies, and their influence on performance, Table~\ref{tab:ablation} reports the percentage change in the MSE metric achieved by each configuration relative to a baseline without the corresponding component. 
A value greater than $0\%$, indicated in green, denotes an improvement in MSE for the corresponding configuration relative to the baseline without it. A value near $\pm1\%$, shown in yellow, suggests that the configuration has a negligible effect on performance. A negative value (i.e., less than $-1\%$), highlighted in red, indicates a performance degradation when the specified configuration is used. For example, in the first row of the table, for the ECL dataset with a prediction horizon of $H=96$, the \textit{Time Embedding} column shows a value of +10.4\% in dark green. This indicates that the best configuration for this task, including \textit{Time Embedding}, achieves a +10.4\% improvement in MSE compared to the best configuration without it.

As previously discussed, we consider each combination of dataset and forecasting horizon as a separate task, reflecting their distinct challenges. Consequently, we perform a task-specific analysis to assess the influence of each architectural change on individual tasks. For more details we refer to Appendix~\ref{appendix:analysis}. 
In the following, we provide a detailed discussion of each architectural variation.

\textbf{Is attention all we need?}
The attention block within the memory module enables dependency modeling and contextual learning across diverse representations, and has demonstrated effectiveness in prior time series forecasting approaches~\cite{zhou2022fedformer, nie2022time, liu2024itransformer}. However, self-attention incurs substantial computational overhead, which must be justified by corresponding performance gains. 
The percentage changes relative to the best configuration, shown in the \textit{Attention} column of Table~\ref{tab:ablation}, indicate a general decrease in performance across most tasks when attention mechanisms were incorporated into the architecture. The degradation is particularly pronounced in the ECL tasks, where performance dropped by up to 9.4\%. For most tasks, the reduction was minor; however, certain cases, such as the Traffic dataset with longer forecast horizons, exhibited a performance improvement of up to 3.2\%. Additionally, we observed a negative correlation of $-0.54$ between the CNN embedding and attention-based configurations, suggesting that both approaches contribute to contextual learning and that employing only one may suffice.

\textbf{Are time-informed patches superior to non-informed patches?}
Time-informed patches incorporate a time embedding to provide the model with temporal context. We analyze whether the inclusion of time embeddings offers any performance gains compared to standard patches without temporal information.
As shown in Table~\ref{tab:ablation}, nearly all tasks benefit from time-informed patches compared to standard patches. The Traffic dataset, in particular, demonstrates considerable performance improvements. In most tasks, the integration of time feature embedding or temporal embedding consistently yields the best results, whereas positional embedding typically enhances performance only when combined with other temporal embeddings.

\textbf{Does the recurrence mechanism of LSTMs enhance performance in patch-based time series forecasting?}
LSTMs facilitate modeling of temporal context. In our approach, we incorporate LSTM layers within the projection module, to capture temporal dependencies between patches before applying the final linear projection. As shown in Table~\ref{tab:ablation}, the ECL and ETTh1 datasets benefit from the inclusion of LSTM layers, while most of the remaining datasets perform better without LSTMs.

\textbf{Is it sometimes better to forget?}
By integrating GLU layers into specific representation embeddings and the memory module, the model effectively filters out irrelevant information, thereby enhancing information flow. As shown in Table~\ref{tab:ablation}, incorporating GLU within the memory module yields a notable performance improvement across almost all tasks. In particular, the ECL and Traffic datasets exhibit substantial gains.

\textbf{How many representatives do we need?}
Most existing work in patch-based time series forecasting uses a fixed patch length to represent the input sequence~\cite{nie2022time, liu2024itransformer}. In our work, use multiple patch extractors in order to receive representations in different abstraction levels, capturing fine-grained features with smaller patches and features like trend with larger patches. The results in Table~\ref{tab:ablation} (Multi Patch) shows superior performance of our architecture when using more than one patch extractors. For most of the tasks, three or four patch extractors resulted in the best performing setups.

\textbf{Do we need the memory stack?}
Stacking memory modules increases model depth, allowing it to capture more complex dependencies in the data. In Table~\ref{tab:ablation}, we compare the performance of a single memory module (\textit{Memory}) against no memory module, as well as stacked memory modules against a single memory module (\textit{Stacked Memory}). The results demonstrate a clear advantage of using a single memory module over none. However, the comparison between a single and multiple memory modules shows task-dependent performance, with no consistent superiority across tasks.

\textbf{Do CNN-based embeddings offer any advantages?}
CNN-based embeddings utilize the feature extraction capabilities of CNNs to enhance representation learning. In this work, we examine three distinct CNN-based embedding strategies, as described in Appendix~\ref{app:representations}. The ablation results presented in Table~\ref{tab:ablation} reveal task-dependent performance: certain datasets—such as ECL, ETTh1, and ETTm2—benefit considerable from CNN-based embeddings, while others, such as Traffic, exhibit performance degradation when these embeddings are employed. Notably, most tasks achieved their best performance using our CNN-based embedding architecture consisting of two convolutional layers, each followed by a max-pooling layer. In contrast, the CNN variant employing two convolutional layers, each followed by a GELU activations rarely yielded competitive results.

\textbf{Is there free lunch?}
No. The “No-Free-Lunch Theorem” implies that no single model performs optimally across all tasks~\cite{wolpert1997no}. Our analysis confirms this by revealing that different tasks impose heterogeneous requirements on the model to achieve accurate forecasting performance. 

\begin{table}[]
\caption{Comparison of architectural variants across tasks. The reported values represent the percentage change in performance relative to a baseline model without the corresponding configuration. Color coding is used to highlight performance impact: values between $-1\%$ and $+1\%$ are shown in yellow, indicating negligible change; values above $+1\%$ are shown in green, indicating performance improvement; values below $-1\%$ are shown in red, indicating degradation.}
\label{tab:ablation}
\centering
\scalebox{.85}{
\begin{tabular}{@{}cccccccccc@{}}
\toprule
\multicolumn{2}{c}{Dataset} & Attention & Time Embedding & GLU & LSTM & Multi Patch & Memory & Multi Memory & CNN Embedding \\ \midrule
\multirow{5}{*}{\rotatebox{90}{\small{ECL}}} & \multicolumn{1}{c|}{96} & \cellcolor[HTML]{EBE8A2} -1.4 & \cellcolor[HTML]{3F8331} +10.4 & \cellcolor[HTML]{B3D4AC} +0.5 & \cellcolor[HTML]{5C9750} +7.6 & \cellcolor[HTML]{B5D6AE} +0.3 & \cellcolor[HTML]{CBE0AE} -0.3 & \cellcolor[HTML]{B6D6AE} +0.3 & \cellcolor[HTML]{8FBB86} +3.4 \\
 & \multicolumn{1}{c|}{192} & \cellcolor[HTML]{B0D2A8} +0.8 & \cellcolor[HTML]{A6CB9E} +1.6 & \cellcolor[HTML]{9FC697} +2.1 & \cellcolor[HTML]{E1EAA8} -0.8 & \cellcolor[HTML]{E1EAA8} -0.8 & \cellcolor[HTML]{A7CC9F} +1.5 & \cellcolor[HTML]{E1EAA8} -0.8 & \cellcolor[HTML]{B0D2A8} +0.8 \\
 & \multicolumn{1}{c|}{336} & \cellcolor[HTML]{E7A180} -5.7 & \cellcolor[HTML]{9CC493} +2.4 & \cellcolor[HTML]{3F8331} +13.5 & \cellcolor[HTML]{A8CCA0} +1.4 & \cellcolor[HTML]{89B67F} +4.0 & \cellcolor[HTML]{A8CCA0} +1.4 & \cellcolor[HTML]{85B47B} +4.3 & \cellcolor[HTML]{E9CD95} -3.0 \\
 & \multicolumn{1}{c|}{720} & \cellcolor[HTML]{E36464} -9.4 & \cellcolor[HTML]{80B176} +4.6 & \cellcolor[HTML]{A1C799} +2.0 & \cellcolor[HTML]{EBDE9D} -2.0 & \cellcolor[HTML]{A1C799} +2.0 & \cellcolor[HTML]{5E9852} +7.5 & \cellcolor[HTML]{A1C799} +2.0 & \cellcolor[HTML]{80B176} +4.6 \\
\multirow{5}{*}{\rotatebox{90}{\small{ETTh1}}} & \multicolumn{1}{c|}{96} & \cellcolor[HTML]{BFDBB1} -0.1 & \cellcolor[HTML]{B8D8B1} +0.1 & \cellcolor[HTML]{B9D8B2} +0.0 & \cellcolor[HTML]{C4DDB0} -0.2 & \cellcolor[HTML]{BFDBB1} -0.1 & \cellcolor[HTML]{B8D8B1} +0.1 & \cellcolor[HTML]{B8D8B1} +0.1 & \cellcolor[HTML]{EAD398} -2.7 \\
 & \multicolumn{1}{c|}{192} & \cellcolor[HTML]{EBEAA2} -1.3 & \cellcolor[HTML]{B0D2A8} +0.8 & \cellcolor[HTML]{BAD9B3} 0.0 & \cellcolor[HTML]{B0D2A8} +0.8 & \cellcolor[HTML]{A4CA9C} +1.7 & \cellcolor[HTML]{EBEAA2} -1.3 & \cellcolor[HTML]{B3D4AC} +0.5 & \cellcolor[HTML]{EBE39F} -1.7 \\
 & \multicolumn{1}{c|}{336} & \cellcolor[HTML]{D9E6AA} -0.6 & \cellcolor[HTML]{9DC595} +2.3 & \cellcolor[HTML]{AFD1A8} +0.8 & \cellcolor[HTML]{B2D3AA} +0.6 & \cellcolor[HTML]{D9E6AA} -0.6 & \cellcolor[HTML]{9DC595} +2.3 & \cellcolor[HTML]{D9E6AA} -0.6 & \cellcolor[HTML]{B2D3AA} +0.6 \\
 & \multicolumn{1}{c|}{720} & \cellcolor[HTML]{E9CC94} -3.1 & \cellcolor[HTML]{8DB983} +3.6 & \cellcolor[HTML]{B4D4AC} +0.5 & \cellcolor[HTML]{93BD8A} +3.1 & \cellcolor[HTML]{408432} +9.8 & \cellcolor[HTML]{EBE7A1} -1.5 & \cellcolor[HTML]{81B177} +4.6 & \cellcolor[HTML]{6CA361} +6.3 \\
\multirow{5}{*}{\rotatebox{90}{\small{ETTh2}}} & \multicolumn{1}{c|}{96} & \cellcolor[HTML]{C9DFAE} -0.3 & \cellcolor[HTML]{C9DFAE} -0.3 & \cellcolor[HTML]{DAE7A9} -0.7 & \cellcolor[HTML]{EBEEA4} -1.1 & \cellcolor[HTML]{ACCFA4} +1.1 & \cellcolor[HTML]{B4D5AD} +0.4 & \cellcolor[HTML]{B0D2A9} +0.8 & \cellcolor[HTML]{D9E6AA} -0.6 \\
 & \multicolumn{1}{c|}{192} & \cellcolor[HTML]{C8DFAE} -0.3 & \cellcolor[HTML]{B9D8B2} +0.0 & \cellcolor[HTML]{B6D6AF} +0.3 & \cellcolor[HTML]{EBE5A0} -1.6 & \cellcolor[HTML]{B5D6AE} +0.3 & \cellcolor[HTML]{ACCFA4} +1.1 & \cellcolor[HTML]{B9D8B2} +0.0 & \cellcolor[HTML]{BBD9B2} -0.0 \\
 & \multicolumn{1}{c|}{336} & \cellcolor[HTML]{ACCFA4} +1.1 & \cellcolor[HTML]{EBE5A0} -1.6 & \cellcolor[HTML]{A6CB9E} +1.6 & \cellcolor[HTML]{EBE49F} -1.7 & \cellcolor[HTML]{98C18F} +2.7 & \cellcolor[HTML]{C9DFAE} -0.3 & \cellcolor[HTML]{A2C89A} +1.9 & \cellcolor[HTML]{E9C893} -3.3 \\
 & \multicolumn{1}{c|}{720} & \cellcolor[HTML]{C7DEAF} -0.3 & \cellcolor[HTML]{C7DEAF} -0.3 & \cellcolor[HTML]{B6D6AE} +0.3 & \cellcolor[HTML]{EBEDA4} -1.1 & \cellcolor[HTML]{B6D6AE} +0.3 & \cellcolor[HTML]{B9D8B2} +0.1 & \cellcolor[HTML]{B6D6AF} +0.3 & \cellcolor[HTML]{EBEDA4} -1.1 \\
\multirow{5}{*}{\rotatebox{90}{\small{ETTm1}}} & \multicolumn{1}{c|}{96} & \cellcolor[HTML]{C3DDB0} -0.2 & \cellcolor[HTML]{C3DDB0} -0.2 & \cellcolor[HTML]{B7D7B0} +0.2 & \cellcolor[HTML]{EBE09E} -1.9 & \cellcolor[HTML]{A9CDA1} +1.4 & \cellcolor[HTML]{9CC494} +2.4 & \cellcolor[HTML]{EBE09E} -1.9 & \cellcolor[HTML]{EBE8A2} -1.4 \\
 & \multicolumn{1}{c|}{192} & \cellcolor[HTML]{E9C390} -3.6 & \cellcolor[HTML]{BCD9B2} -0.0 & \cellcolor[HTML]{BCD9B2} -0.0 & \cellcolor[HTML]{E1EAA8} -0.8 & \cellcolor[HTML]{B0D2A8} +0.8 & \cellcolor[HTML]{B0D2A8} +0.8 & \cellcolor[HTML]{E1EAA8} -0.8 & \cellcolor[HTML]{B8D8B1} +0.1 \\
 & \multicolumn{1}{c|}{336} & \cellcolor[HTML]{B7D7B0} +0.2 & \cellcolor[HTML]{A9CDA1} +1.3 & \cellcolor[HTML]{C5DEAF} -0.2 & \cellcolor[HTML]{E4EBA7} -0.8 & \cellcolor[HTML]{A5CA9D} +1.7 & \cellcolor[HTML]{D3E4AB} -0.5 & \cellcolor[HTML]{B6D6AF} +0.3 & \cellcolor[HTML]{B3D4AC} +0.5 \\
 & \multicolumn{1}{c|}{720} & \cellcolor[HTML]{C5DDAF} -0.2 & \cellcolor[HTML]{C5DDAF} -0.2 & \cellcolor[HTML]{B8D8B1} +0.1 & \cellcolor[HTML]{C5DDAF} -0.2 & \cellcolor[HTML]{A5CA9D} +1.7 & \cellcolor[HTML]{B8D7B1} +0.1 & \cellcolor[HTML]{B8D8B1} +0.1 & \cellcolor[HTML]{B8D8B1} +0.1 \\
\multirow{5}{*}{\rotatebox{90}{\small{ETTm2}}} & \multicolumn{1}{c|}{96} & \cellcolor[HTML]{EACF96} -2.9 & \cellcolor[HTML]{83B379} +4.4 & \cellcolor[HTML]{ACCFA5} +1.1 & \cellcolor[HTML]{E8BD8D} -4.0 & \cellcolor[HTML]{88B67E} +4.0 & \cellcolor[HTML]{8DBA84} +3.6 & \cellcolor[HTML]{EBE6A0} -1.5 & \cellcolor[HTML]{A7CB9E} +1.5 \\
 & \multicolumn{1}{c|}{192} & \cellcolor[HTML]{AED0A6} +1.0 & \cellcolor[HTML]{97C08E} +2.8 & \cellcolor[HTML]{A4CA9C} +1.7 & \cellcolor[HTML]{EBE5A0} -1.6 & \cellcolor[HTML]{A2C899} +1.9 & \cellcolor[HTML]{97C08E} +2.8 & \cellcolor[HTML]{EADD9C} -2.0 & \cellcolor[HTML]{AED0A6} +1.0 \\
 & \multicolumn{1}{c|}{336} & \cellcolor[HTML]{D3E4AC} -0.5 & \cellcolor[HTML]{9EC596} +2.2 & \cellcolor[HTML]{A8CDA0} +1.4 & \cellcolor[HTML]{EBE8A2} -1.4 & \cellcolor[HTML]{B3D4AC} +0.5 & \cellcolor[HTML]{A0C798} +2.0 & \cellcolor[HTML]{BEDAB1} -0.1 & \cellcolor[HTML]{B3D4AC} +0.5 \\
 & \multicolumn{1}{c|}{720} & \cellcolor[HTML]{EAD397} -2.7 & \cellcolor[HTML]{95BF8C} +3.0 & \cellcolor[HTML]{C0DBB1} -0.1 & \cellcolor[HTML]{E8BA8C} -4.2 & \cellcolor[HTML]{8FBA85} +3.5 & \cellcolor[HTML]{B3D4AC} +0.5 & \cellcolor[HTML]{8CB982} +3.7 & \cellcolor[HTML]{9DC594} +2.3 \\
\multirow{5}{*}{\rotatebox{90}{\small{Traffic}}} & \multicolumn{1}{c|}{96} & \cellcolor[HTML]{B4D5AD} +0.4 & \cellcolor[HTML]{3F8331} +13.7 & \cellcolor[HTML]{9CC493} +2.4 & \cellcolor[HTML]{B3D4AC} +0.5 & \cellcolor[HTML]{3F8331} +14.0 & \cellcolor[HTML]{92BD88} +3.2 & \cellcolor[HTML]{CFE2AC} -0.4 & \cellcolor[HTML]{EADE9D} -2.0 \\
 & \multicolumn{1}{c|}{192} & \cellcolor[HTML]{EBE7A1} -1.4 & \cellcolor[HTML]{3F8331} +10.7 & \cellcolor[HTML]{98C18F} +2.7 & \cellcolor[HTML]{9BC392} +2.5 & \cellcolor[HTML]{A8CCA0} +1.4 & \cellcolor[HTML]{EAD699} -2.5 & \cellcolor[HTML]{ADD0A5} +1.0 & \cellcolor[HTML]{EADE9D} -2.0 \\
 & \multicolumn{1}{c|}{336} & \cellcolor[HTML]{92BD88} +3.2 & \cellcolor[HTML]{4A8A3C} +9.1 & \cellcolor[HTML]{78AB6D} +5.3 & \cellcolor[HTML]{EAD89A} -2.4 & \cellcolor[HTML]{8CB982} +3.7 & \cellcolor[HTML]{EBE9A2} -1.3 & \cellcolor[HTML]{8CB982} +3.7 & \cellcolor[HTML]{EAD89A} -2.4 \\
 & \multicolumn{1}{c|}{720} & \cellcolor[HTML]{9FC697} +2.1 & \cellcolor[HTML]{458737} +9.5 & \cellcolor[HTML]{83B279} +4.4 & \cellcolor[HTML]{EADC9C} -2.1 & \cellcolor[HTML]{82B278} +4.5 & \cellcolor[HTML]{D4E4AB} -0.5 & \cellcolor[HTML]{99C290} +2.6 & \cellcolor[HTML]{E8B58A} -4.5 \\
\multirow{5}{*}{\rotatebox{90}{\small{Weather}}} & \multicolumn{1}{c|}{96} & \cellcolor[HTML]{D2E3AC} -0.5 & \cellcolor[HTML]{B2D4AB} +0.6 & \cellcolor[HTML]{B2D4AB} +0.6 & \cellcolor[HTML]{EAD99A} -2.3 & \cellcolor[HTML]{BDDAB2} -0.1 & \cellcolor[HTML]{7AAC70} +5.1 & \cellcolor[HTML]{EAD99A} -2.3 & \cellcolor[HTML]{BDDAB2} -0.1 \\
 & \multicolumn{1}{c|}{192} & \cellcolor[HTML]{DBE7A9} -0.7 & \cellcolor[HTML]{B1D3AA} +0.7 & \cellcolor[HTML]{B0D2A8} +0.8 & \cellcolor[HTML]{EBE29F} -1.8 & \cellcolor[HTML]{B6D6AF} +0.3 & \cellcolor[HTML]{97C08E} +2.8 & \cellcolor[HTML]{DCE8A9} -0.7 & \cellcolor[HTML]{DBE7A9} -0.7 \\
 & \multicolumn{1}{c|}{336} & \cellcolor[HTML]{EBEBA3} -1.2 & \cellcolor[HTML]{A8CCA0} +1.4 & \cellcolor[HTML]{B9D8B2} +0.0 & \cellcolor[HTML]{EBE9A2} -1.3 & \cellcolor[HTML]{D1E3AC} -0.5 & \cellcolor[HTML]{D1E3AC} -0.5 & \cellcolor[HTML]{E5EBA6} -0.9 & \cellcolor[HTML]{B4D4AC} +0.5 \\
 & \multicolumn{1}{c|}{720} & \cellcolor[HTML]{D2E3AC} -0.5 & \cellcolor[HTML]{A4CA9C} +1.7 & \cellcolor[HTML]{E0E9A8} -0.8 & \cellcolor[HTML]{EBE19E} -1.9 & \cellcolor[HTML]{AED1A7} +0.9 & \cellcolor[HTML]{9BC392} +2.5 & \cellcolor[HTML]{D2E3AC} -0.5 & \cellcolor[HTML]{D2E3AC} -0.5 \\
\end{tabular}
}
\end{table}
\section{Limitations}\label{sec:limitations}
Although REP-Net achieved strong performance on the benchmark datasets used in our experiments, several potential limitations remain as outlined below.

    \textbf{Benchmark datasets}: While our empirical study employed well-established benchmark datasets, it focused exclusively on long-term forecasting, leaving short-term forecasting performance unexamined. In addition, existing benchmark datasets exhibit several limitations, as outlined by Bergmeir.~\citep{limitations_TSFC}.
    
    \textbf{Impact of outliers}: While the model demonstrated stable performance across most datasets, we observed a significant degradation in the MSE metric on the Traffic dataset, attributable to a single feature (840) that contains outliers in the test set. Further investigation is necessary to ensure a more robust computation of REP-Net.
    
    \textbf{Model Analysis}: Despite substantial efforts to analyze the impact of various architectural modifications on model performance, our analysis was restricted to a fixed set of architectural variations and did not isolate the side effects of different hyperparameters. Further experiments are necessary to investigate the influence of specific hyperparameters on model performance within particular architectural configurations.
    
    \textbf{Complexity and runtime}: While the current system demonstrates promising resource efficiency and low inference latency, our comparison with other methods revealed instances of superior inference speed and reduced memory usage. These findings warrant further investigation to reduce the computational complexity of REP-Net.

\section{Conclusion}
We propose REP-Net, a novel architecture for time series forecasting that decomposes the forecasting pipeline into three distinct stages: Representation, Memory, and Projection. Within each stage, we investigate a diverse set of architectural variants, including time-aware patching strategies, multi-level abstractions in the representation phase, various embedding techniques, the integration of self-attention and gated linear units (GLUs) for information enrichment, and LSTM mechanisms over patch sequences. REP-Net is explicitly designed to maintain low computational complexity and a minimal memory footprint. Empirical results demonstrate that REP-Net matches or outperforms SOTA methods across several established time series forecasting benchmarks, while reducing the number of trainable parameters and inference time.
We conducted a comprehensive analysis of architectural variations on each time series forecasting task individually, revealing key insights into the effectiveness of specific components within the architecture.
The findings in this paper highlight the significance of task-specific architectural designs that adapt to the unique properties of both the dataset and the forecasting horizon to enable high-quality predictions. Architectural variations, such as incorporating GLU layers and time-informed patches within the forecasting pipeline, prove to be highly effective for most tasks involved in our study.

\newpage

\bibliographystyle{unsrtnat}
\bibliography{references}  


\newpage
\appendix
\section{Related Work}\label{appendix:related_work}

\textbf{Transformer-based models.} Since its inception in 2017, Transformer~\citep{vaswani2017attention} and its numerous derivatives steadily gained traction and are now a well-established approach to time series modeling. Approaches like Informer~\citep{zhou2021informer} addressed the quadratic complexity of vanilla attention, replacing it with $O(n \log n)$ or linear attention mechanisms. Autoformer~\cite{wu2021autoformer} replaced vanilla attention with autocorrelation and series decomposition blocks. Shortly after, Zhou et al.~\cite{zhou2022fedformer} followed a similar path using frequency transformations and a different flavor of decomposition. Around the same time, PatchTST~\citep{nie2022time} combined the Transformer encoder with subseries-level patches as input encoding to increase efficiency while demonstrating strong modeling capacity. While PatchTST processes each channel of multivariate TS independently, Crossformer~\citep{zhang2023crossformer} captures both temporal and cross-channel dependencies. To this end, the model unravels the input TS into two dimensions and features a novel attention layer to learn both types of dependencies efficiently. Pathformer~\citep{chen2024pathformer} is a multiscale transformer with adaptive dual attention to capture temporal dependencies between TS segments of varying granularity. Lastly, Liu et al.~\cite{liu2024itransformer} proposed the iTransformer that applies the attention and linear layers on the inverted dimensions to capture multivariate correlations in time series.

\textbf{Self-supervised pretraining.} Splitting the training process into pretraining and fine-tuning allows TS models to learn universal representations that can be later utilized for different downstream tasks~\citep{jiang2022transferability}. Examples include SimMTM~\citep{dong2023simmtm} and its successor HiMTM~\citep{zhao2024himtm}, which employ a convolution- and self-attention-based encoder architecture. SimMTM was introduced as a pre-training framework for masked time series modeling that recovers masked time points by the weighted aggregation of multiple neighbors outside the data manifold. HiMTM persues hierarchical multi-scale masked time series modeling with self-distillation, incorporating cross-scale attention fine-tuning. In contrast, CoST~\citep{woo2022cost} learns disentangled feature representations by discriminating the trend and seasonal components. Lee et al.~\cite{lee2024learning} recently presented PITS, which leverages a combination of patch-based autoencoding and contrastive learning. 

\textbf{Foundation models.} Similar to self-supervised pretraining, time series foundation models learn universal representations of TS and use them for different downstream tasks~\citep{bommasani2021opportunities}. However, they are more powerful in that they pretrain on a cross-domain database to generalize across individual target datasets. In recent years, various approaches have been proposed, including TF-C~\citep{zhang2022self}, TimesNet~\citep{wu2022timesnet}, FPT~\citep{zhou2023one}, Lag-Llama~\citep{rasul2023lag}, MOMENT~\citep{goswami2024moment}, MOIRAI~\citep{woo2024unified}, and TimesFM~\citep{das2024decoder}. TimesFM and FPT are Transformer-based models. TF-C employs an embedding stage based on time-frequency-consistency and contrastive learning. Further, TimesNet analyses temporal variations in the 1D input sequence by unfolding it into two dimensions along multiple periods observed over the time axis. MOIRAI follows a patch-based approach with a masked encoder architecture.

\textbf{Patch-based models.} Patching is a form of input encoding that divides the time series into subsequences, which can be either overlapping or non-overlapping~\citep{nie2022time, zhang2023crossformer, zhou2023one, das2024decoder, lee2024learning, chen2024pathformer, liu2024itransformer, goswami2024moment, woo2024unified}. In the basic form, identical-sized patches are sliced from the input TS and fed as tokens to the model~\citep{nie2022time}. Pathformer's multiscale division splits the TS into different temporal resolutions using patches of various, dynamically chosen sizes. Crossformer computes more complex patches, encoding both temporal and cross-channel dependencies. iTransformer~\citep{liu2024itransformer} takes the idea to the extreme, operating on patches covering an entire channel of the input TS each.

\textbf{Few/zero-shot learning.} Few-shot learning refers to the capability of a model to generalize from the data domain it is (pre-)trained on to a new target domain using just a few (zero-shot learning: none) target-training instances~\citep{zhou2023one, rasul2023lag, das2024decoder, lee2024learning, woo2024unified}. Lag-Llama is based on a decoder-only Transformer architecture that uses lags as covariates to process univariate TS. Moreover, Lag-Llama is pretrained on a large corpus of multidomain TS data, while FPT utilizes a pretrained language model like BERT~\citep{devlin2018bert} as basis.

\textbf{Linear time series modeling.} With the rise of more and more complex deep neural models for time series forecasting like transformers, issues regarding interpretability, scalability, and complexity became increasingly pressing. As a consequence, some research focused on simpler and light-weight architectures based on linear mapping again. DLinear~\cite{zeng2023transformers}, for instance, is a combination of a decomposition scheme used in \cite{wu2021autoformer} with linear layers. First, the input time series is decomposed into trend and season. Then, each component is separately feed into a one layer network and the outputs added to a final prediction. A similar approach is RLinear~\cite{li2023revisiting}, which uses a linear layer combined with reversible normalization to demonstrate the effectiveness of linear mapping for learning periodicity in long-term
time series forecasting. TSMixer, proposed by Chen et al.~\cite{chen2023tsmixer}, extended the concept to multi-layer perceptrons (MLP). It utilizes linear layers performing mixing operations along the time and feature axis to extract information effectively. Similarly, TimeMixer~\cite{wang2024timemixer} is an MLP-based architecture that combines series decomposition from~\cite{wu2021autoformer} with linear layers arranged in two mixing blocks to extract information from different TS scales first, then aggregate the disentangled representation into a coherent prediction. Two additional methods, CycleNet~\cite{lin2024cyclenet} and TimeCNN~\cite{hu2024timecnn}, also revolve around an MLP backbone and linear projections, but are extended by modeling of periodic patterns and cross-variable interactions, respectively. CycleNet introduced residual cycle forecasting, which leverages learnable recurrent cycles and the prediction of residual components. TimeCNN innovates a timepoint-indepentent CNN module, where each time point of the input sequence has its independent convolution kernel. All of these methods achieved results comparable or exceeding to considerably more complex models.

\textbf{Structured state space models.} Recently, Gu et al.~\cite{gu2022efficiently} introduced another approach to sequence modeling based on structured state space models (S4). S4s were first introduced for language modeling and demonstrate state-of-the-art performance, especially on long sequence tasks. While initial variants are convolution-based, subsequent approaches, such as Mamba~\cite{gu2024mamba}, switch to recurrent designs, further improving performance and simplifying the model architecture at the same time. Ahamed et al.~\cite{ahamed2024timemachine} transferred the concept to the time series forecasting task. They proposed TimeMachine, which is built around four Mamba architectures, to address long-term, multi-variate forecasting, performing favorably compared to transformer-, CNN-, and MLP-based baselines.  

\textbf{Delimitation.} Our work does not fit well into any of these areas. We do not present a single architectural design, but rather a pattern based on the combination and configuration of three subsequent components: Representation, Memory, and Projection. The goal is to understand how the configuration of each component and their relationships impact the performance of the model. Additionally, our analysis includes the role of attention, the impact of LSTMs on projection capacity, and comparison of patch extractors. The similarities with previous work are more fine-grained. For instance, we experiment with self-attention popularized by transformers and linear encodings akin to linear time series models. Furthermore, our representation component is based on patching, where we test different extractor functions. The memory component may also be stacked, which as a concept is popular with previous architectures such as transformers, linear TS models, or TimeMachine.

\section{Datasets}\label{appendix:datasets}

Below, we provide more details on the datasets used in our experiments. Please find a detailed overview of all employed benchmark datasets in Table~\ref{tab:datasets}.

\textbf{Electricity Load Diagram (ECL)}: 
The Electricity dataset, available at UCI~\citep{dua2017uci}, contains electricity consumption data measured in kilowatt-hours (kWh). It includes data from 370 clients collected every 15 minutes for 48 months, starting from January 2011 to December 2014. 

\textbf{Electricity Transformer Temperature (ETT)}:
The ETT dataset comprises data collected from electricity transformers over a time period from July 1, 2016, to June 26, 2018. ETT consists of 4 subsets, where ETTh1 and ETTh2 contain records with hourly resolution, while ETTm1 and ETTm2 are recorded every 15 minutes. In total, ETT includes 69,680 data points without any missing values. Each record contains seven features, including oil temperature and six different types of external power load features~\citep{zhou2021informer}. 


\textbf{Traffic}:
The Traffic dataset~\cite{traffic_dataset} comprehensively represents occupancy rates within California's freeway infrastructure. It is derived from 862 integrated sensor systems, which monitored vehicular capacity usage from July 2016 through 2018, culminating in 17,544 hourly records per sensor. Each sensor is interpreted as a channel of multivariate TS. The dataset was partitioned such that the initial 70\% constituted the training subset, the subsequent 10\% formed the validation subset, and the final 20\% comprised the test subset.

\textbf{Weather}: The weather dataset contains the recordings of 21 meteorological factors, such as temperature, humidity, and air pressure, collected every 10 minutes from the weather station of the Max Planck Biogeochemistry Institute in Jena, Germany in 2020~\citep{wu2021autoformer}.

\begin{table}[ht]
\caption{Details of the used benchmark datasets. The assignment to train, validation, or test follows the established procedure~\citep{wu2021autoformer}.}
\label{tab:datasets}
\centering
\scalebox{.85}{
\begin{tabular}{@{}lllll@{}}
\toprule
Dataset & Channels & Size (train / val / test) & Frequency & Information \\ \midrule
ECL & 321 & 18317 / 2633 / 5261 & Hourly & Electricity \\
ETTm1,ETTm2 & 7 & 34465 / 11521 / 11521 & 15min & Electricity \\
ETTh1,ETTh2 & 7 & 8545 / 2881 / 2881 & Hourly & Electricity \\
Traffic & 862 & 12280 / 1754 / 3508 & Hourly & Traffic \\
Weather & 21 & 36792 / 5271 / 10540 & 10min & Weather \\
\bottomrule
\end{tabular}
}
\end{table}

\section{Methodology}
\label{appendix:methodology}

\paragraph{Hyperparameters}\label{app:hyperparameters}

We present below the complete set of hyperparameters considered in our hyperparameter study. The study was conducted using randomized search.

\begin{itemize}
    \item Attention heads $h \in \{4, 8, 16, 32\}$
    \item Amount of stacked memory modules $N \in \{0, 1, 2, 3, 4, 5, 8, 10\}$, where $N=0$ means the model does not include a memory module. 
    \item Feature embedding size $e_f \in \{4, 8, 16, 32, 64, 128\}$
    \item Time embedding size $e_t \in \{8, 16\}$
    \item Patch extractors $K \in \{1,2,3,4,5\}$
    \item Patch extractor cover sizes $\{3, 5, 10, 15, 20, 48, 64\}$
    \item Time embedding method: We evaluate the use of time features ($timeF$), temporal embeddings ($tempEmb$), positional embeddings ($posEmb$), all possible combinations of these three, as well as a baseline with no embeddings.
    \item Embedding techniques (see below)
    \item Dropout $\text{dropout} \in \{0.25, 0.33, 0.5, 0.66, 0.9\}$.
    \item Attention or no attention, which cause the entire attention block to be removed from the architecture.
    \item Amount of stacked LSTM layer $R \in \{0, 1, 2, 3\}$, where $R = 0$ means the model does not include an LSTM layer.
    \item The linear layer in the first block of the memory module can be configured either to operate across all feature embeddings jointly or to process each feature embedding independently.
    
\end{itemize}

\paragraph{Embedding Techniques}\label{app:representations}

\begin{itemize}
    \item A single linear layer ($E_1$).
    \item Two linear layers ($E_1$).
    \item Two linear layers with a GELU activation in between ($E_1$).
    \item Two linear layers with a GELU activation in between, followed by a Gated Linear Unit (GLU) layer ($E_2$).
    \item A single CNN layer with kernel size three, followed by a linear layer ($E_3$).
    \item Two CNN layers with kernel size three, each followed by a GELU activation, and a linear layer at the end ($E_3$).
    \item Two CNN layers with kernel size three, each followed by a max pooling layer with kernel size three, and a linear layer at the end ($E_3$).
\end{itemize}

\section{Results}
\label{appendix:results}

We provide detailed results of the forecasting experiments that supplement Table~\ref{tab:results} in Table~\ref{app:tab:results}. Specifically, we report individual performance metrics for each dataset and forecasting horizon, along with the average performance across all horizons. 

\begin{table}[]
\centering
\caption{Performance comparison for the multivariate forecasting task with prediction horizons $H \in \{96, 192, 336, 720\}$ and fixed lookback window $T = 96$. AVG shows the averaged result over all prediction horizons per dataset and model. Bold/underline indicate best/second. Best includes a margin of 1\%.}
\label{app:tab:results}

\scalebox{.49}{
\begin{tabular}{@{}ccllllllllllllllllllllll@{}}
\toprule
\multicolumn{2}{c}{Dataset} & \multicolumn{2}{c}{\textbf{REP-Net}} & \multicolumn{2}{c}{\begin{tabular}[c]{@{}c@{}}CycleNet \\ (NeurIPS 2024)\end{tabular}} & \multicolumn{2}{c}{\begin{tabular}[c]{@{}c@{}}TimeMixer \\ (ICLR 2024)\end{tabular}} & \multicolumn{2}{c}{\begin{tabular}[c]{@{}c@{}}TimeMachine \\ (ECAI 2024)\end{tabular}} & \multicolumn{2}{c}{TimeCNN} & \multicolumn{2}{c}{TSRM} & \multicolumn{2}{c}{\begin{tabular}[c]{@{}c@{}}PathFormer\\ (ICLR 2024)\end{tabular}} & \multicolumn{2}{c}{\begin{tabular}[c]{@{}c@{}}iTransformer\\ (ICLR 2024)\end{tabular}} & \multicolumn{2}{c}{\begin{tabular}[c]{@{}c@{}}PatchTST\\ (ICLR 2023)\end{tabular}} & \multicolumn{2}{c}{RLinear} & \multicolumn{2}{c}{DLinear} \\ \midrule
\multicolumn{2}{c|}{} & \multicolumn{1}{c}{MSE} & \multicolumn{1}{c|}{MAE} & \multicolumn{1}{c}{MSE} & \multicolumn{1}{c}{MAE} & \multicolumn{1}{c}{MSE} & \multicolumn{1}{c}{MAE} & \multicolumn{1}{c}{MSE} & \multicolumn{1}{c}{MAE} & \multicolumn{1}{c}{MSE} & \multicolumn{1}{c}{MAE} & \multicolumn{1}{c}{MSE} & \multicolumn{1}{c}{MAE} & \multicolumn{1}{c}{MSE} & \multicolumn{1}{c}{MAE} & \multicolumn{1}{c}{MSE} & \multicolumn{1}{c}{MAE} & \multicolumn{1}{c}{MSE} & \multicolumn{1}{c}{MAE} & \multicolumn{1}{c}{MSE} & \multicolumn{1}{c}{MAE} & \multicolumn{1}{c}{MSE} & \multicolumn{1}{c}{MAE} \\
\multirow{5}{*}{\rotatebox{90}{\small{ECL}}} & \multicolumn{1}{c|}{96} & \underline{0.140} & \multicolumn{1}{l|}{0.240} & \textbf{0.136} & \textbf{0.229} & 0.153 & 0.247 & 0.142 & \underline{0.236} & \underline{0.140} & \underline{0.236} & 0.148 & 0.245 & 0.145 & \underline{0.236} & 0.148 & 0.240 & 0.166 & 0.252 & 0.201 & 0.281 & 0.197 & 0.282 \\
 & \multicolumn{1}{c|}{192} & 0.158 & \multicolumn{1}{l|}{0.252} & \textbf{0.152} & \textbf{0.244} & 0.166 & 0.256 & 0.158 & \underline{0.250} & \underline{0.156} & \underline{0.250} & 0.176 & 0.262 & 0.167 & 0.256 & 0.162 & 0.253 & 0.174 & 0.260 & 0.201 & 0.283 & 0.196 & 0.285 \\
 & \multicolumn{1}{c|}{336} & \textbf{0.169} & \multicolumn{1}{l|}{0.274} & \textbf{0.169} & \textbf{0.264} & 0.185 & 0.277 & 0.172 & 0.268 & 0.172 & \underline{0.267} & 0.175 & 0.274 & 0.186 & 0.275 & 0.178 & 0.269 & 0.191 & 0.278 & 0.215 & 0.298 & 0.209 & 0.301 \\
 & \multicolumn{1}{c|}{720} & \textbf{0.193} & \multicolumn{1}{l|}{\textbf{0.296}} & 0.210 & \textbf{0.296} & 0.225 & 0.310 & 0.207 & \textbf{0.296} & 0.212 & 0.304 & \underline{0.202} & 0.299 & 0.231 & 0.309 & 0.225 & 0.317 & 0.230 & 0.311 & 0.257 & 0.331 & 0.245 & 0.333 \\
 & \multicolumn{1}{c|}{AVG} & \textbf{0.165} & \multicolumn{1}{l|}{0.266} & \underline{0.167} & \textbf{0.258} & 0.182 & 0.273 & 0.170 & \underline{0.263} & 0.170 & 0.264 & 0.175 & 0.270 & 0.182 & 0.269 & 0.178 & 0.270 & 0.190 & 0.275 & 0.219 & 0.298 & 0.212 & 0.300 \\
\multirow{5}{*}{\rotatebox{90}{\small{Weather}}} & \multicolumn{1}{c|}{96} & \textbf{0.149} & \multicolumn{1}{l|}{\underline{0.196}} & 0.158 & 0.203 & 0.163 & 0.209 & 0.164 & 0.208 & 0.167 & 0.210 & \underline{0.153} & 0.200 & 0.156 & \textbf{0.192} & 0.174 & 0.214 & 0.177 & 0.218 & 0.192 & 0.232 & 0.196 & 0.255 \\
 & \multicolumn{1}{c|}{192} & \textbf{0.198} & \multicolumn{1}{l|}{\underline{0.243}} & 0.207 & 0.247 & 0.208 & 0.250 & 0.211 & 0.250 & 0.215 & 0.254 & \underline{0.202} & 0.245 & 0.206 & \textbf{0.240} & 0.221 & 0.254 & 0.223 & 0.257 & 0.240 & 0.271 & 0.237 & 0.296 \\
 & \multicolumn{1}{c|}{336} & 0.256 & \multicolumn{1}{l|}{0.286} & 0.262 & 0.289 & \textbf{0.251} & 0.287 & 0.256 & 0.290 & 0.272 & 0.296 & 0.261 & \underline{0.285} & \underline{0.254} & \textbf{0.282} & 0.278 & 0.296 & 0.277 & 0.297 & 0.292 & 0.307 & 0.283 & 0.335 \\
 & \multicolumn{1}{c|}{720} & \textbf{0.332} & \multicolumn{1}{l|}{\textbf{0.335}} & 0.344 & 0.344 & \underline{0.339} & 0.341 & 0.342 & 0.343 & 0.355 & 0.351 & 0.341 & 0.339 & 0.340 & \textbf{0.335} & 0.358 & 0.347 & 0.365 & 0.367 & 0.364 & 0.353 & 0.345 & 0.381 \\
 & \multicolumn{1}{c|}{AVG} & \textbf{0.234} & \multicolumn{1}{l|}{\textbf{0.263}} & 0.243 & 0.271 & 0.240 & 0.272 & 0.243 & 0.273 & 0.252 & 0.278 & \underline{0.239} & 0.267 & \underline{0.239} & \textbf{0.263} & 0.258 & 0.278 & 0.261 & 0.285 & 0.272 & 0.291 & 0.265 & 0.317 \\
\multirow{5}{*}{\rotatebox{90}{\small{Traffic}}} & \multicolumn{1}{c|}{96} & 0.412 & \multicolumn{1}{l|}{\underline{0.262}} & 0.458 & 0.296 & 0.462 & 0.285 & 0.397 & 0.268 & \textbf{0.377} & \textbf{0.257} & 0.540 & 0.321 & 0.479 & 0.283 & \underline{0.395} & 0.268 & 0.446 & 0.286 & 0.649 & 0.389 & 0.650 & 0.396 \\
 & \multicolumn{1}{c|}{192} & 0.425 & \multicolumn{1}{l|}{\underline{0.274}} & 0.457 & 0.294 & 0.473 & 0.296 & \underline{0.417} & \underline{0.274} & \textbf{0.398} & \textbf{0.267} & 0.524 & 0.312 & 0.484 & 0.292 & \underline{0.417} & 0.276 & 0.453 & 0.285 & 0.601 & 0.366 & 0.598 & 0.370 \\
 & \multicolumn{1}{c|}{336} & 0.439 & \multicolumn{1}{l|}{\underline{0.280}} & 0.470 & 0.299 & 0.498 & 0.296 & \underline{0.433} & 0.281 & \textbf{0.416} & \textbf{0.274} & 0.544 & 0.326 & 0.503 & 0.299 & \underline{0.433} & 0.283 & 0.467 & 0.291 & 0.609 & 0.369 & 0.605 & 0.373 \\
 & \multicolumn{1}{c|}{720} & 0.473 & \multicolumn{1}{l|}{0.301} & 0.502 & 0.314 & 0.506 & 0.313 & \underline{0.467} & \underline{0.300} & \textbf{0.450} & \textbf{0.293} & 0.571 & 0.336 & 0.537 & 0.322 & \underline{0.467} & 0.302 & 0.500 & 0.309 & 0.647 & 0.387 & 0.645 & 0.394 \\
 & \multicolumn{1}{c|}{AVG} & 0.437 & \multicolumn{1}{l|}{\underline{0.279}} & 0.472 & 0.301 & 0.485 & 0.298 & 0.429 & 0.281 & \textbf{0.410} & \textbf{0.273} & 0.545 & 0.324 & 0.501 & 0.299 & \underline{0.428} & 0.282 & 0.467 & 0.293 & 0.627 & 0.378 & 0.625 & 0.383 \\
\multirow{5}{*}{\rotatebox{90}{\small{ETTh1}}} & \multicolumn{1}{c|}{96} & \textbf{0.364} & \multicolumn{1}{l|}{\underline{0.391}} & 0.375 & 0.395 & 0.375 & 0.400 & \textbf{0.364} & \textbf{0.387} & 0.374 & 0.393 & 0.377 & 0.396 & 0.382 & 0.400 & 0.386 & 0.405 & 0.395 & 0.407 & 0.386 & 0.395 & 0.386 & 0.400 \\
 & \multicolumn{1}{c|}{192} & \textbf{0.409} & \multicolumn{1}{l|}{\textbf{0.416}} & 0.426 & \textbf{0.416} & 0.429 & 0.421 & \underline{0.415} & \textbf{0.416} & 0.430 & 0.424 & 0.427 & 0.424 & 0.440 & 0.427 & 0.441 & 0.436 & 0.446 & 0.434 & 0.437 & 0.424 & 0.437 & 0.432 \\
 & \multicolumn{1}{c|}{336} & \underline{0.445} & \multicolumn{1}{l|}{0.438} & 0.464 & 0.439 & 0.484 & 0.458 & \textbf{0.429} & \textbf{0.421} & 0.476 & 0.449 & 0.461 & 0.442 & 0.454 & \underline{0.432} & 0.487 & 0.458 & 0.489 & 0.457 & 0.479 & 0.446 & 0.481 & 0.459 \\
 & \multicolumn{1}{c|}{720} & \textbf{0.449} & \multicolumn{1}{l|}{\textbf{0.452}} & 0.461 & 0.460 & 0.498 & 0.482 & \underline{0.458} & \textbf{0.452} & 0.485 & 0.476 & 0.474 & 0.459 & 0.479 & 0.461 & 0.503 & 0.491 & 0.488 & 0.477 & 0.481 & 0.470 & 0.519 & 0.516 \\
 & \multicolumn{1}{c|}{AVG} & \textbf{0.417} & \multicolumn{1}{l|}{\underline{0.425}} & 0.432 & 0.428 & 0.447 & 0.440 & \textbf{0.417} & \textbf{0.419} & 0.441 & 0.436 & 0.435 & 0.430 & 0.439 & 0.430 & 0.454 & 0.448 & 0.455 & 0.444 & 0.446 & 0.434 & 0.456 & 0.452 \\
\multirow{5}{*}{\rotatebox{90}{\small{ETTh2}}} & \multicolumn{1}{c|}{96} & \textbf{0.275} & \multicolumn{1}{l|}{\textbf{0.330}} & 0.285 & 0.335 & 0.289 & 0.341 & \textbf{0.275} & 0.334 & 0.293 & 0.344 & 0.285 & \textbf{0.330} & 0.279 & \textbf{0.330} & 0.297 & 0.349 & 0.292 & 0.342 & 0.288 & 0.338 & 0.333 & 0.387 \\
 & \multicolumn{1}{c|}{192} & \textbf{0.349} & \multicolumn{1}{l|}{\textbf{0.379}} & 0.372 & 0.396 & 0.372 & 0.392 & \textbf{0.349} & \textbf{0.379} & 0.372 & 0.394 & 0.368 & 0.384 & \textbf{0.349} & \textbf{0.379} & 0.380 & 0.400 & 0.378 & 0.393 & 0.374 & 0.390 & 0.477 & 0.476 \\
 & \multicolumn{1}{c|}{336} & 0.390 & \multicolumn{1}{l|}{0.423} & 0.421 & 0.433 & 0.386 & 0.414 & \textbf{0.340} & \textbf{0.381} & 0.416 & 0.429 & 0.414 & 0.429 & \underline{0.348} & \textbf{0.381} & 0.428 & 0.432 & 0.420 & 0.429 & 0.415 & 0.426 & 0.594 & 0.541 \\
 & \multicolumn{1}{c|}{720} & \textbf{0.391} & \multicolumn{1}{l|}{\textbf{0.424}} & 0.450 & 0.458 & 0.412 & 0.434 & 0.411 & 0.433 & 0.423 & 0.442 & 0.417 & 0.442 & \underline{0.398} & \textbf{0.424} & 0.427 & 0.445 & 0.423 & 0.442 & 0.420 & 0.440 & 0.831 & 0.657 \\
 & \multicolumn{1}{c|}{AVG} & 0.352 & \multicolumn{1}{l|}{0.389} & 0.382 & 0.406 & 0.365 & 0.395 & \textbf{0.344} & \textbf{0.379} & 0.376 & 0.402 & 0.371 & 0.397 & \textbf{0.344} & \textbf{0.379} & 0.383 & 0.407 & 0.378 & 0.402 & 0.374 & 0.399 & 0.559 & 0.515 \\
\multirow{5}{*}{\rotatebox{90}{\small{ETTm1}}} & \multicolumn{1}{c|}{96} & \textbf{0.303} & \multicolumn{1}{l|}{\underline{0.351}} & 0.319 & 0.360 & 0.320 & 0.357 & 0.317 & 0.355 & 0.315 & \underline{0.351} & \underline{0.314} & 0.352 & 0.316 & \textbf{0.346} & 0.334 & 0.368 & 0.325 & 0.361 & 0.355 & 0.376 & 0.345 & 0.372 \\
 & \multicolumn{1}{c|}{192} & \textbf{0.347} & \multicolumn{1}{l|}{\textbf{0.370}} & 0.360 & 0.381 & 0.361 & 0.381 & \underline{0.357} & 0.378 & 0.360 & 0.377 & 0.363 & 0.382 & 0.366 & \textbf{0.370} & 0.377 & 0.391 & 0.362 & 0.383 & 0.391 & 0.392 & 0.380 & 0.389 \\
 & \multicolumn{1}{c|}{336} & \textbf{0.375} & \multicolumn{1}{l|}{\textbf{0.394}} & 0.389 & 0.403 & 0.390 & 0.404 & \underline{0.379} & 0.399 & 0.396 & 0.401 & 0.393 & 0.401 & 0.386 & \textbf{0.394} & 0.426 & 0.420 & 0.392 & 0.402 & 0.424 & 0.415 & 0.413 & 0.413 \\
 & \multicolumn{1}{c|}{720} & \textbf{0.440} & \multicolumn{1}{l|}{0.440} & 0.447 & 0.441 & 0.454 & 0.441 & \underline{0.445} & \textbf{0.432} & 0.460 & 0.439 & 0.448 & \textbf{0.432} & 0.460 & \textbf{0.432} & 0.491 & 0.459 & 0.456 & \textbf{0.432} & 0.487 & 0.450 & 0.474 & 0.453 \\
 & \multicolumn{1}{c|}{AVG} & \textbf{0.366} & \multicolumn{1}{l|}{\underline{0.390}} & 0.379 & 0.396 & 0.381 & 0.396 & \underline{0.375} & 0.392 & 0.383 & 0.392 & 0.380 & 0.393 & 0.382 & \textbf{0.386} & 0.407 & 0.410 & 0.384 & 0.396 & 0.414 & 0.408 & 0.403 & 0.407 \\
\multirow{5}{*}{\rotatebox{90}{\small{ETTm2}}} & \multicolumn{1}{c|}{96} & \textbf{0.163} & \multicolumn{1}{l|}{\textbf{0.246}} & \textbf{0.163} & \textbf{0.246} & 0.175 & 0.258 & 0.175 & 0.256 & 0.176 & 0.258 & 0.169 & 0.253 & 0.170 & \textbf{0.246} & 0.188 & 0.264 & 0.178 & 0.260 & 0.182 & 0.265 & 0.193 & 0.292 \\
 & \multicolumn{1}{c|}{192} & \textbf{0.227} & \multicolumn{1}{l|}{\underline{0.294}} & \textbf{0.227} & \textbf{0.290} & 0.237 & 0.299 & 0.239 & 0.299 & 0.239 & 0.300 & 0.236 & 0.297 & 0.238 & 0.295 & 0.250 & 0.309 & 0.247 & 0.305 & 0.246 & 0.304 & 0.284 & 0.362 \\
 & \multicolumn{1}{c|}{336} & \underline{0.287} & \multicolumn{1}{l|}{0.332} & \textbf{0.284} & \textbf{0.327} & 0.298 & 0.340 & \underline{0.287} & 0.332 & 0.299 & 0.340 & 0.298 & 0.397 & 0.293 & \underline{0.331} & 0.311 & 0.348 & 0.302 & 0.341 & 0.307 & 0.342 & 0.369 & 0.427 \\
 & \multicolumn{1}{c|}{720} & 0.377 & \multicolumn{1}{l|}{0.392} & 0.389 & 0.391 & 0.391 & 0.396 & \underline{0.371} & \underline{0.385} & 0.404 & 0.401 & \textbf{0.276} & \textbf{0.321} & 0.390 & 0.389 & 0.412 & 0.407 & 0.407 & 0.401 & 0.407 & 0.398 & 0.554 & 0.522 \\
 & \multicolumn{1}{c|}{AVG} & \underline{0.264} & \multicolumn{1}{l|}{\textbf{0.314}} & 0.266 & \textbf{0.314} & 0.275 & 0.323 & 0.268 & 0.318 & 0.280 & 0.325 & \textbf{0.245} & \textbf{0.314} & 0.273 & \textbf{0.314} & 0.290 & 0.332 & 0.284 & 0.327 & 0.286 & 0.327 & 0.350 & 0.401 \\ \cmidrule(l){2-24} 
\end{tabular}
}
\end{table}

\section{Model Analysis}
\label{appendix:analysis}

In the following, we extend the analysis of our model, point by point with more details to further support our theses.

\subsection{Is Attention All We Need?}

\begin{figure}
    \centering
    \includegraphics[width=0.93\linewidth]{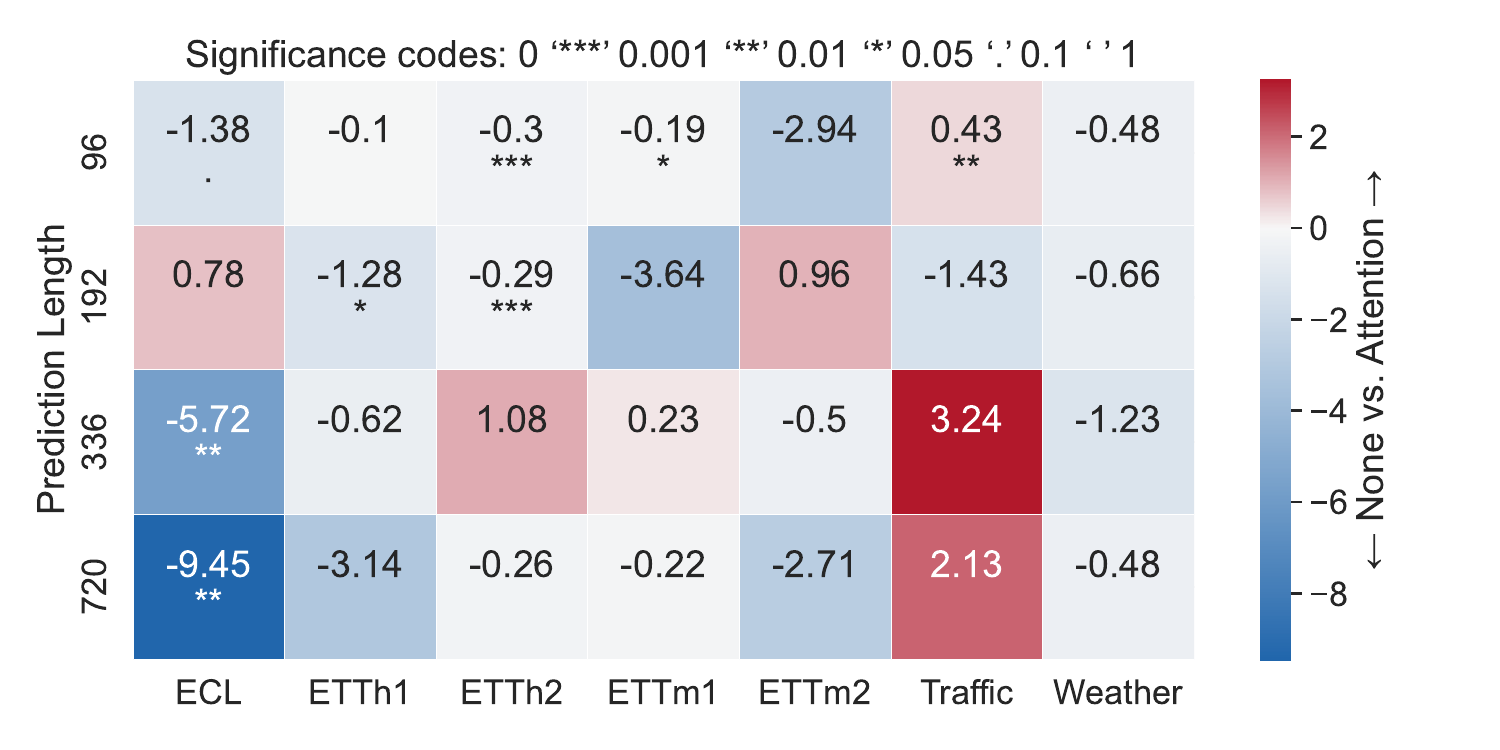}
    \caption{Comparison of models with attention and without. Each row represents one of the four prediction lengths and each column a dataset. Red shading/positive values indicate that models with attention are better, blue shading/negative values that models without attention are better. The reported values are percentages, representing relative difference in performance. More *s means higher statistical significance, as determined by a two-sided paired t-test.}
    \label{fig:attention_comp}
\end{figure}

Across the 28 tasks, the use of attention generally did not lead to statistically significant improvements in forecasting performance as illustrated in Figure~\ref{fig:attention_comp}. In seven cases, models with attention performed better than those without (positive red tiles), but only one of these differences was statistically significant. Conversely, models without attention outperformed those with attention in 21 cases, with six of these differences reaching statistical significance. Notably, there are instances---such as at prediction length 720 on the ECL dataset---where the MSE with attention is significantly higher, suggesting that attention may sometimes introduce noise or cause overfitting. Overall, in six out of 28 cases, models without attention showed significantly better performance, while in a majority of 21 cases, the differences were negligible, and in only one case (length 720, Traffic) attention caused a significant improvement. Given that self-attention mechanisms come with substantial computational overhead, their use should be justified by consistent and meaningful gains.

\subsection{Are Time-Informed Patches Superior to Non-Informed Patches?}

\begin{figure}
    \centering
    \includegraphics[width=0.93\linewidth]{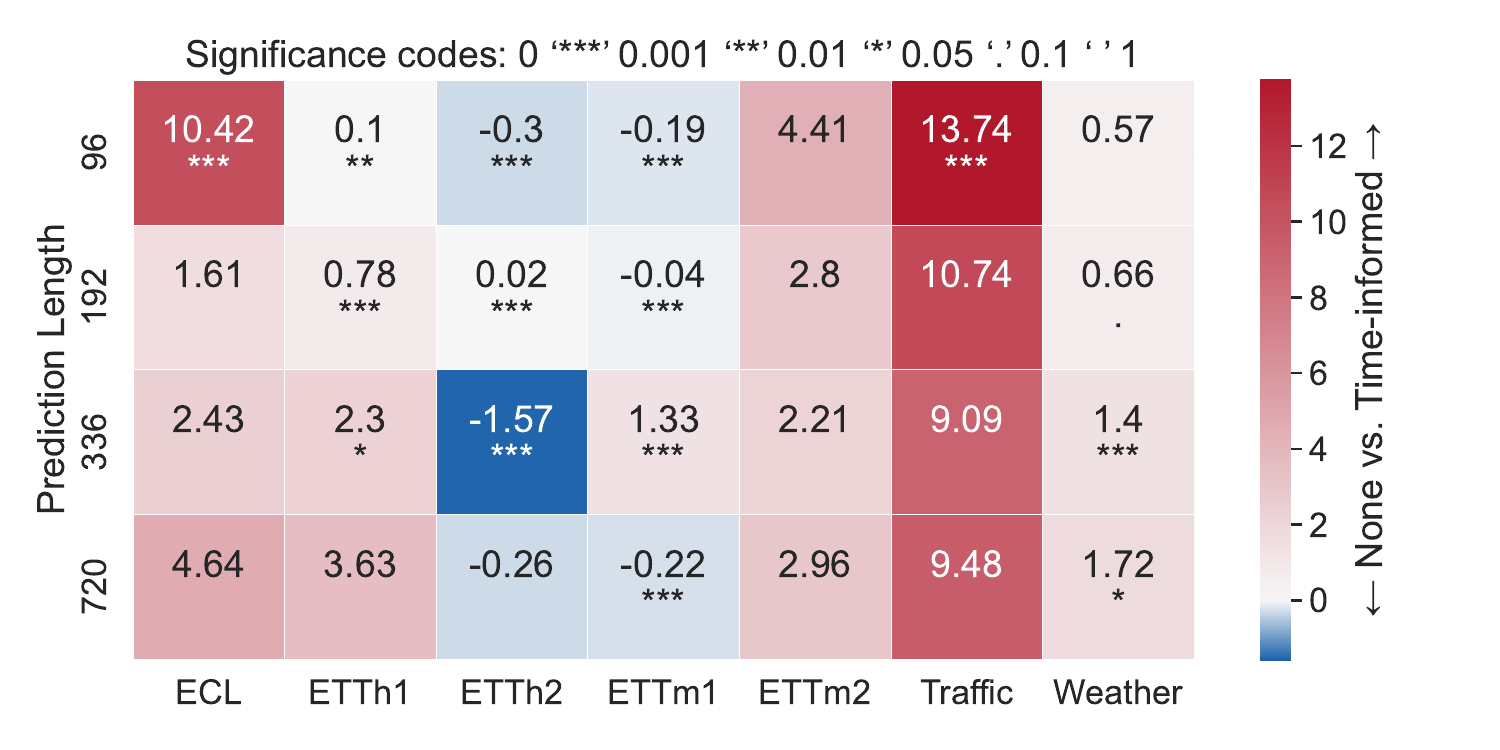}
    \caption{Comparison of models with time-informed patches and non-informed patches. Each row represents one of the four prediction lengths and each column a dataset. Red shading/positive values indicate that models with time-informed patches are better, blue shading/negative values that models without are better. The reported values are percentages, representing relative difference in performance. More *s means higher statistical significance, as determined by a two-sided paired t-test.}
    \label{fig:time_comp}
\end{figure}

Across the 28 tasks (illustrated in Figure~\ref{fig:time_comp}), models using time-informed patches generally outperformed those without them. In 22 cases, time-informed patches led to better forecasting performance (positive tiles), with nine of these differences being statistically significant. The most notable improvement was observed on the Traffic dataset with a prediction length of 96. In contrast, models without time-informed patches performed better in only six cases, with five of those differences reaching statistical significance. These results suggest that, while time-informed patches do not guarantee improvements in every case, they tend to provide consistent and sometimes substantial gains, especially in more challenging forecasting scenarios.

\subsection{Does the Recurrence Mechanism of LSTMs Enhance Performance in Patch-based Time Series Forecasting?}

\begin{figure}
    \centering
    \includegraphics[width=0.93\linewidth]{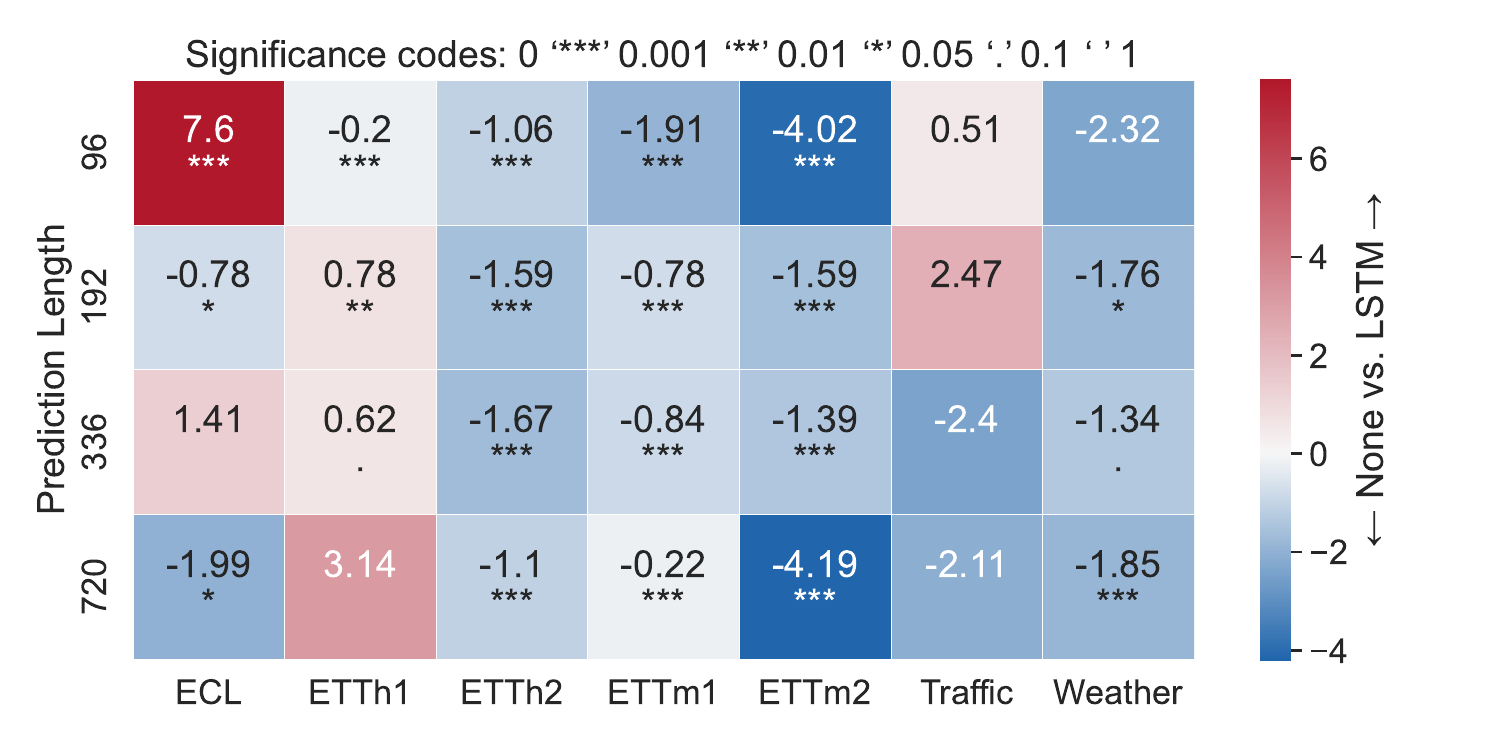}
    \caption{Comparison of models with LSTM and without LSTM. Each row represents one of the four prediction lengths and each column a dataset. Red shading/positive values indicate that models with LSTM are better, blue shading/negative values that models without are better. The reported values are percentages, representing relative difference in performance. More *s means higher statistical significance, as determined by a two-sided paired t-test.}
    \label{fig:lstm_comp}
\end{figure}

Out of the 28 tasks evaluated, the inclusion of LSTM layers led to better performance in seven cases (positive tiles) as shown in Figure~\ref{fig:lstm_comp}, though only two of these showed statistically significant improvements. In contrast, models without LSTM layers outperformed their counterparts in the majority of tasks, with 17 of these differences being statistically significant. Notably, the ECL and ETTh1 datasets showed clear benefits from the use of LSTM layers, indicating that their effectiveness may be dataset-dependent. Overall, these results suggest that while LSTM layers can enhance performance in select scenarios, their broader applicability in time series forecasting tasks appears limited and should be evaluated carefully against their computational cost and the characteristics of the data.

\subsection{Is It Sometimes Better to Forget?}
\begin{figure}
    \centering
    \includegraphics[width=0.93\linewidth]{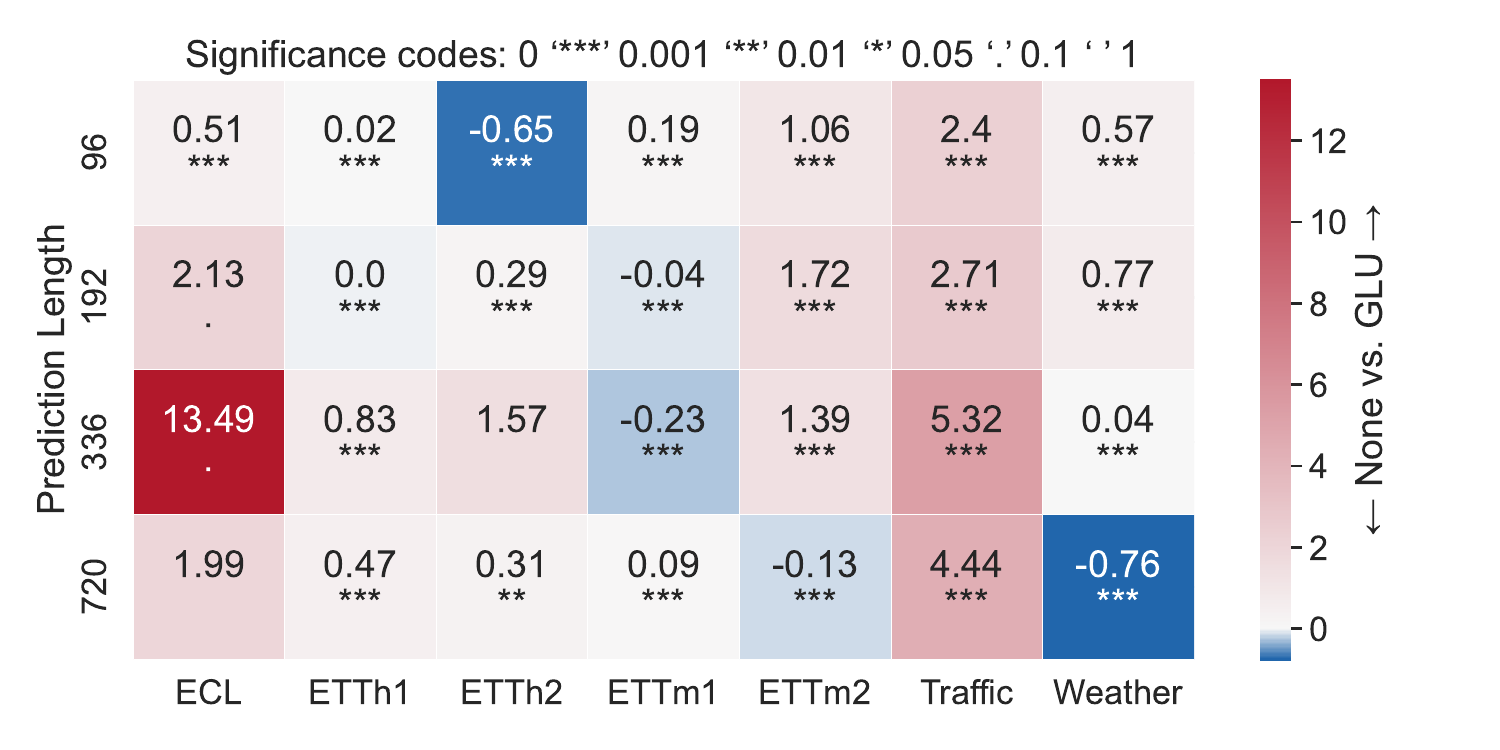}
    \caption{Comparison of models with GLU and without GLU. Each row represents one of the four prediction lengths and each column a dataset. Red shading/positive values indicate that models using GLU are better, blue shading/negative values that models without are better. The reported values are percentages, representing relative difference in performance. More *s means higher statistical significance, as determined by a two-sided paired t-test.}
    \label{fig:glu_comp}
\end{figure}

Models incorporated GLU yielded a better performance in 23 out of 28 tasks (positive tiles) as shown in Figure~\ref{fig:glu_comp}, with 19 of these improvements reaching statistical significance. In contrast, models without GLU performed better in only five cases, though all five were also statistically significant. The ECL and Traffic datasets, in particular, exhibit substantial performance gains from using GLU, suggesting that the gating mechanism effectively filters out irrelevant information and enhances the flow of relevant signals through the model. 

\subsection{How Many Representatives Do We Need?}
\begin{figure}
    \centering
    \includegraphics[width=0.93\linewidth]{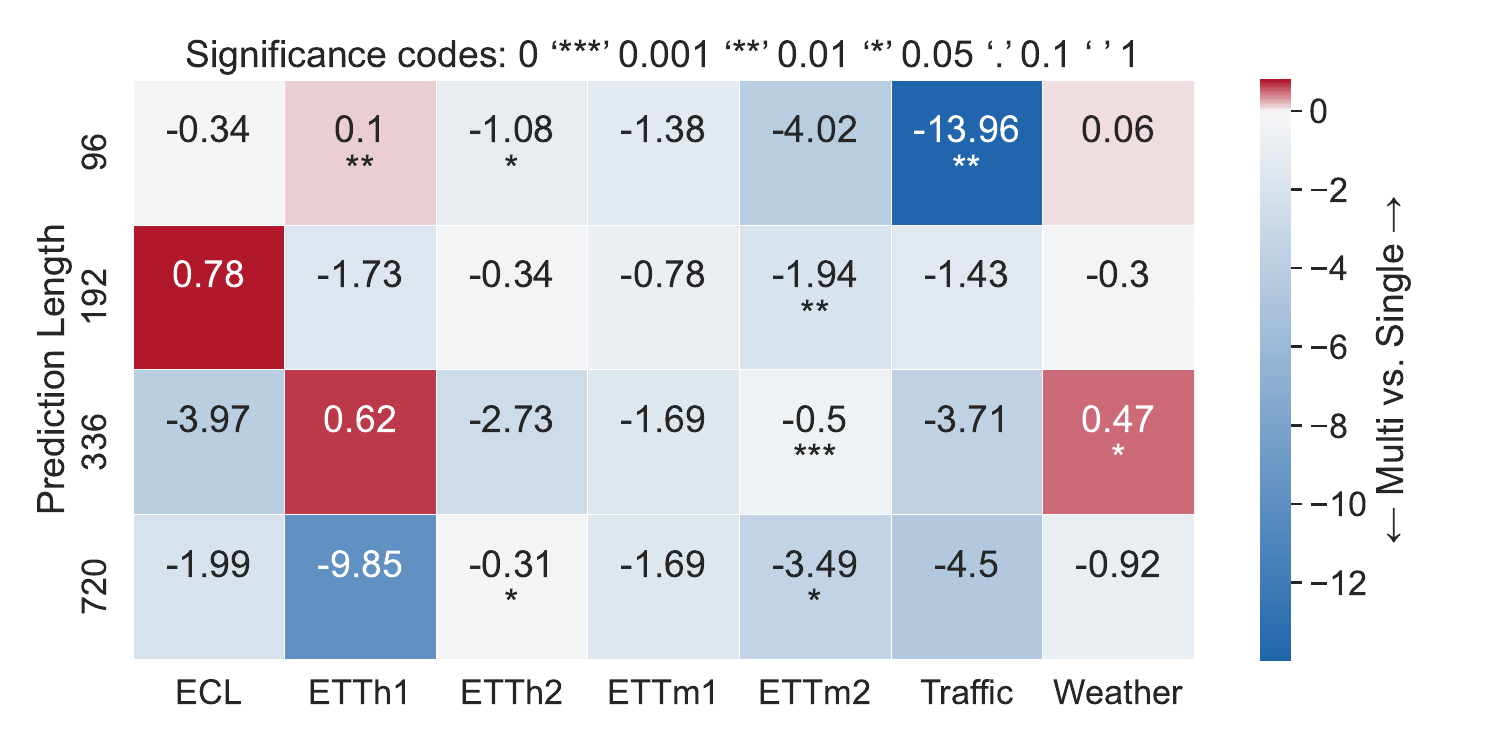}
    \caption{Comparison of models with a single patch extractor and multiple patch extractors. Each row represents one of the four prediction lengths and each column a dataset. Red shading/positive values indicate that models using one patch extractor are better, blue shading/negative values that models using multiple are better. The reported values are percentages, representing relative difference in performance. More *s means higher statistical significance, as determined by a two-sided paired t-test.}
    \label{fig:patch_comp}
\end{figure}

When comparing single-patch to multi-patch configurations as illustrated in Figure~\ref{fig:patch_comp}, models using multiple patches outperformed single-patch models in 23 out of 28 tasks (negative tiles), with six of those differences being statistically significant. In contrast, single-patch models were better in only five cases, with just two showing significance. The most notable performance gain from using multiple patches was observed on the Traffic dataset with a prediction length of 96. These results suggest that leveraging multiple patches can improve forecasting accuracy by capturing a richer representation of temporal dependencies.

\subsection{Do we need the memory stack?}
\begin{figure}
    \centering
    \includegraphics[width=0.93\linewidth]{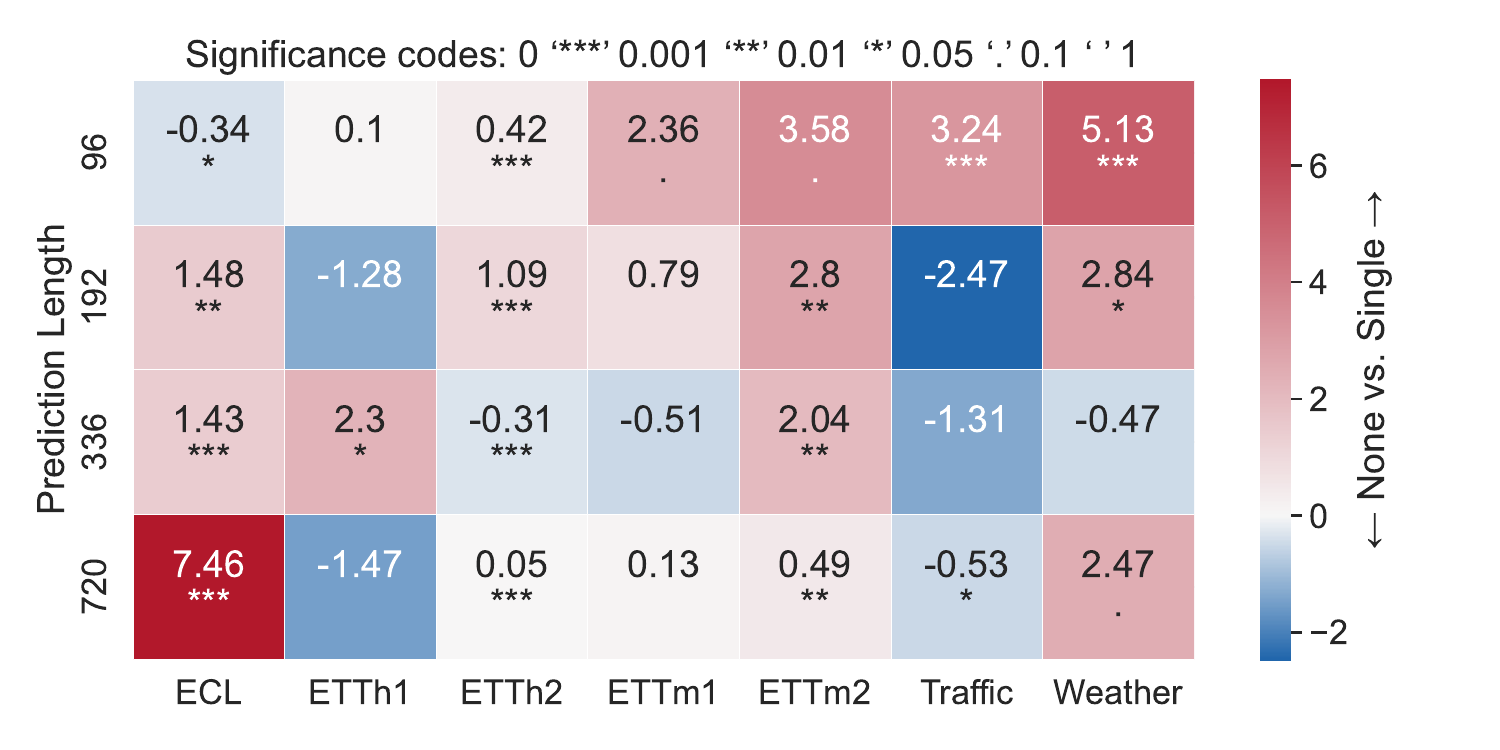}
    \caption{Comparison of models with a single memory module and no memory module. Each row represents one of the four prediction lengths and each column a dataset. Red shading/positive values indicate that models using  a single memory module, blue shading/negative values that models without are better. The reported values are percentages, representing relative difference in performance. More *s means higher statistical significance, as determined by a two-sided paired t-test.}
    \label{fig:memory_single}
\end{figure}

\begin{figure}
    \centering
    \includegraphics[width=0.93\linewidth]{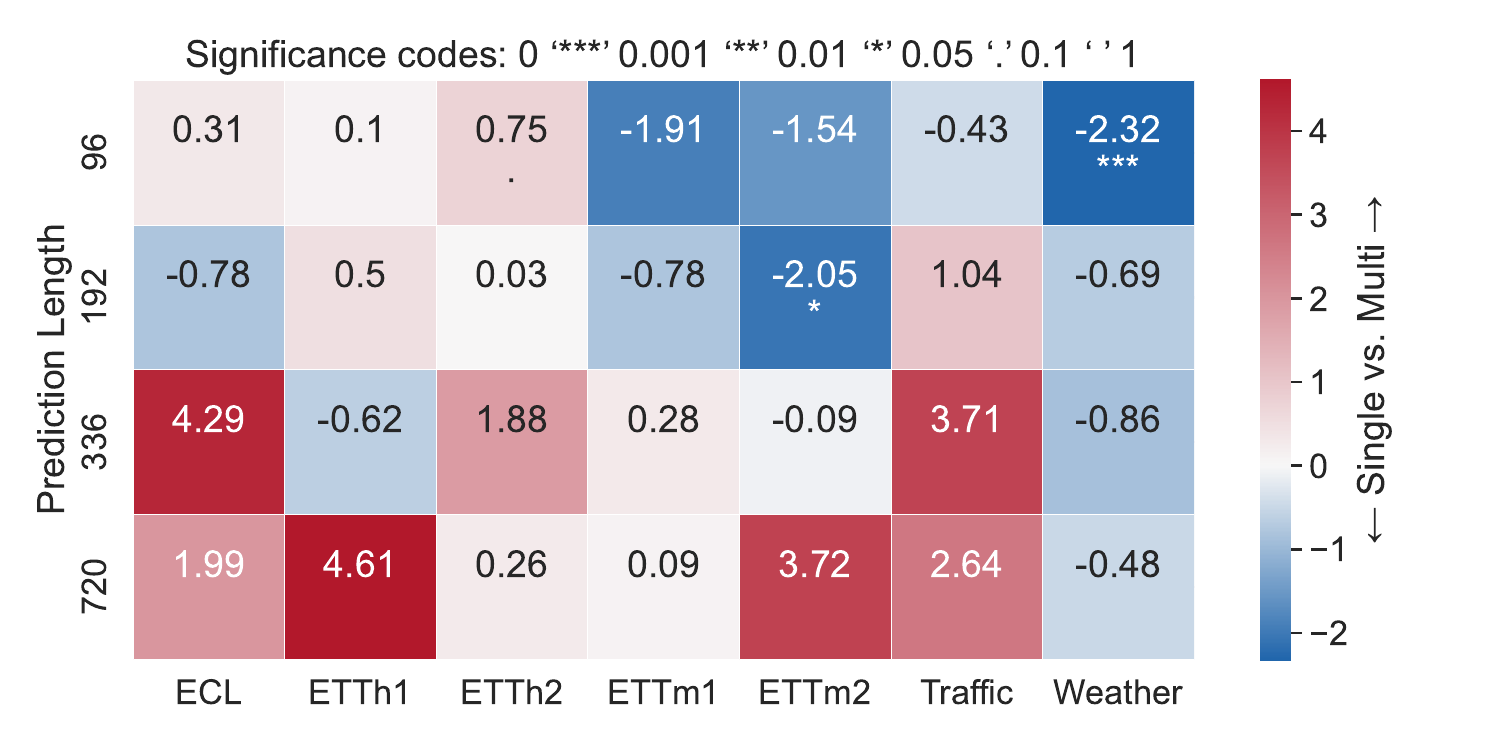}
    \caption{Comparison of models with a single memory module and multiple memory modules. Each row represents one of the four prediction lengths and each column a dataset. Red shading/positive values indicate that models using multiple memory modules, blue shading/negative values that models using a single one are better. The reported values are percentages, representing relative difference in performance. More *s means higher statistical significance, as determined by a two-sided paired t-test.}
    \label{fig:memory_multi}
\end{figure}

When comparing models with a single memory module to those without any memory module (see Figure~\ref{fig:memory_single}), the single-module variant performed better in 19 out of 28 tasks (positive tiles), with 13 of these differences being statistically significant. In contrast, models without a memory module outperformed in nine tasks, with three showing statistical significance. A separate comparison between models with a single memory module and those with multiple memory modules showed more mixed results (refer to Figure~\ref{fig:memory_multi}): the single-module configuration performed better in 12 tasks (negative tiles), with two statistically significant improvements, while the multi-module setup outperformed in 16 tasks, though none of these differences reached statistical significance.

\subsection{Do CNN-based embeddings offer any advantages?}
\begin{figure}
    \centering
    \includegraphics[width=0.93\linewidth]{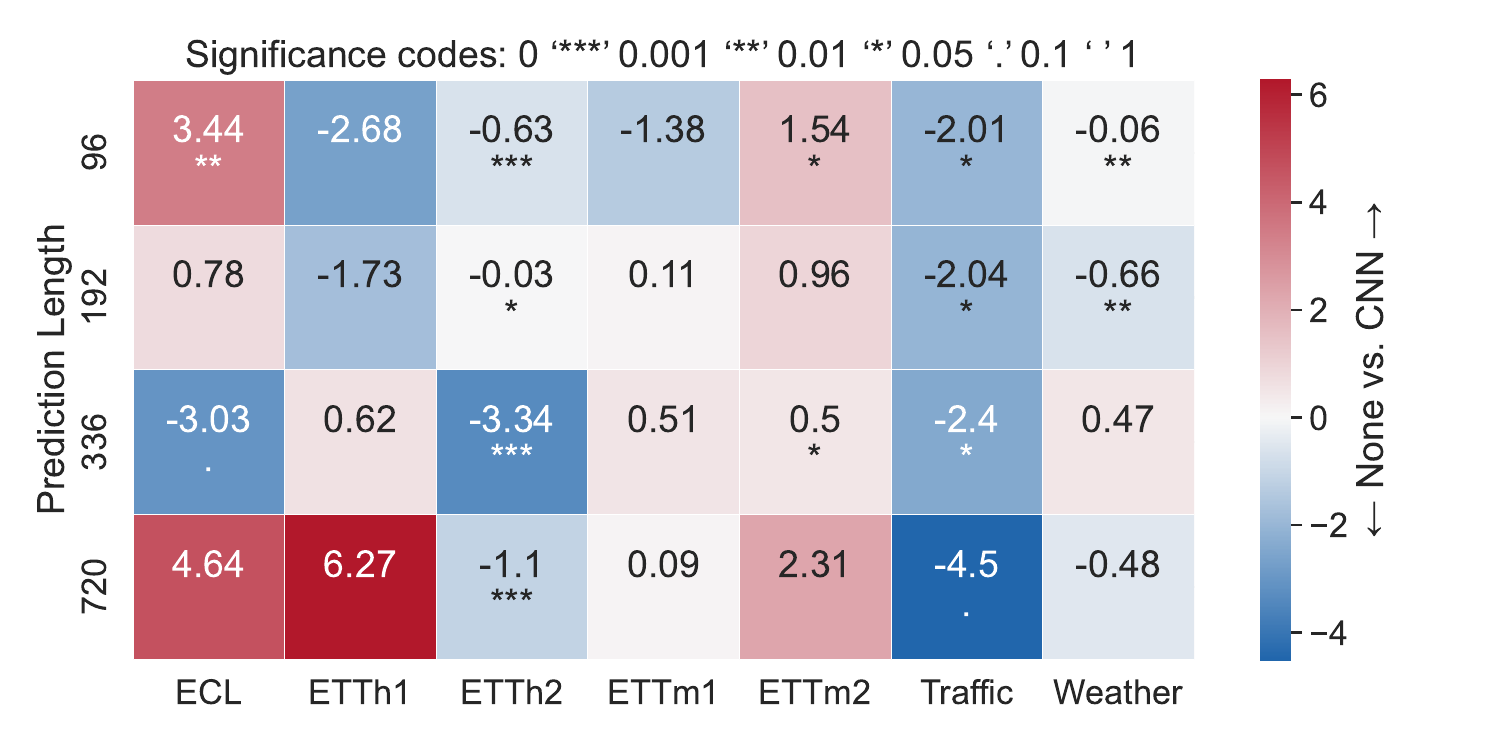}
    \caption{Comparison of models with CNN and without CNN. Each row represents one of the four prediction lengths and each column a dataset. Red shading/positive values indicate that models using CNN-based embeddings, blue shading/negative values that models without are better. The reported values are percentages, representing relative difference in performance. More *s means higher statistical significance, as determined by a two-sided paired t-test.}
    \label{fig:cnn}
\end{figure}

In the case of CNN integration, models showed improved performance in 13 tasks (positive tiles) as shown in Figure~\ref{fig:cnn}, with three of these improvements being statistically significant. Conversely, excluding CNN resulted in performance drops in 15 tasks, nine of which were statistically significant. The most notable gain from using CNN was observed on the ETTm2 dataset, highlighting its potential effectiveness in specific scenarios.

\end{document}